\documentclass[final,1p,times]{tempcls}

\usepackage{amsmath,amssymb,amsfonts}
\usepackage{amsopn}
\usepackage{graphicx}
\usepackage{subfigure}
\usepackage{multirow}

\journal{}

\begin{document}

\begin{frontmatter}



\title{Face Recognition via Globality-Locality Preserving Projections}


\author[a]{Sheng Huang\corref{cor1}}
\author[a,b]{Dan Yang}
\author[c]{Fei Yang}
\author[b,d]{Yongxin Ge}
\author[b,d]{Xiaohong Zhang}
\author[e]{Jiwen Lu}
\address[a]{College of Computer Science at Chongqing University, Chonqing, 400044, China}
\address[b]{School of Software Engineering at Chongqing University Chonqing, 400044, China}
\address[c]{Department of Computer Science at Rutgers University, Piscataway, NJ, 08854, USA}
\address[d]{Ministry of Education Key Laboratory of Dependable Service Computing in Cyber Physical Society at Chongqing University Chonqing, 400044, China}
\address[e]{Advanced Digital Sciences Center (ADSC), 138632, Singapore}
\cortext[cor1]{Corresponding author (Sheng Huang):huangsheng@cqu.edu.cn}

\begin{abstract}
We present an improved Locality Preserving Projections (LPP) method, named \emph{Gloablity-Locality Preserving Projections} (GLPP), to preserve both the global and local geometric structures of data. In our approach, an additional constraint of the geometry of classes is imposed to the objective function of conventional LPP for respecting some more global manifold structures. Moreover, we formulate a two-dimensional extension of GLPP (2D-GLPP) as an example to show how to extend GLPP with some other statistical techniques. We apply our works to face recognition on four popular face databases, namely ORL, Yale, FERET and LFW-A databases, and extensive experimental results demonstrate that the considered global manifold information can significantly improve the performance of LPP and the proposed face recognition methods outperform the state-of-the-arts.
\end{abstract}


\end{frontmatter}


\section{Introduction}
Face recognition is considered as one of the most challenging tasks in computer vision, and has been experienced a vivid enthusiasm during the past decades. Due to various facial expressions, poses and illuminations, it is a challenging problem to extract effective and discriminative features from faces for recognition. To solve this issue, many classical approaches have been proposed. Among them, the most popular kind of face recognition method may be the appearance-based approach. These methods can be divided into two categories, \emph{nonlinear models} and \emph{linear models}. The representative linear methods include Principal Component Analysis (PCA) \cite{pca}, Linear Discriminant Analysis (LDA) \cite{lda}, Nonnegative Matrix Factorization (NMF) \cite{nmf}, Locality Preservation Projection (LPP) \cite{lpp,lap}, and their two-dimensional extensions \cite{2dpca,2dlda,2dlpp,2ddlpp}. While the representative nonlinear methods include Isomap \cite{isomap}, locally linear embedding (LLE) \cite{lle}, and kernel methods \cite{kernel,kernels,klda}. Generally speaking, Linear approaches are simpler and faster than nonlinear approaches. In the following paragraphs, we briefly review the representative linear approaches.

	Among all the linear appearance based methods, Principle Component Analysis (PCA) \cite{pca} may be the most classical unsupervised method. It aims at finding a linear projection for preserving the global variability of data via maximizing the variation of the projected samples. PCA is a efficient method for dimension reduction. However it ignores the class-specific information which is suitable for classification. To solve this problem, many researchers try to develop different kinds of algorithms to combine the class-specific information with PCA \cite{cipca,dbs}.

As another classical linear method, Linear Discriminant Analysis (LDA) \cite{lda} discriminates the data by maximizing the between-class scatter matrix and minimizing the within-class scatter matrix simultaneously. Thus, the homogeneous data points can be projected much closer while the inhomogeneous data points will be projected further. It is generally considered that LDA is superior to PCA for classification. However, LDA suffer the Small Size Sample (SSS) problem. To alleviate this problem, extensive approaches have been proposed in the literatures \cite{ssslda,mmm,mms}. However, the fundamental limitation still remains unsolved in theory.
	
Nonnegative Matrix Factorization (NMF) \cite{nmf}, is a recent approach for extracting a part-based linear representation, which has received great attention and has been widely applied in face recognition area. NMF attempts to decompose a large non-negative matrix into the product of two small non-negative matrices and produces a part-based representation since only additive combination of basis is allowed. Some previous studies have indicated that the mechanism of NMF is very similar to the visual perception mechanism of human brain \cite{nmf}. NMF-based methods are developed rapidly \cite{tnmf,onmf,nmff,gnmf}. However, training these NMF-based methods is computationally expensive compared with other linear methods.

	Some studies show that the face images reside on a nonlinear submanifold \cite{isomap,lle}. These studies boost many manifold learning methods for solving face recognition and other computer vision tasks \cite{lpp,lap,lape,nmf,ge,rlpda,mmd}. Among these manifold learning algorithms, Locality Preservation Projection (LPP) may be the most influential manifold learning algorithm for face recognition and dimensionality reduction. LPP provides a way of projection via constructing an adjacency weighting matrix of data for preserving local manifold structures. Since the objective function is linear, it can be efficiently computed. Although LPP has been applied in many domains and achieves promising results, it seems to still have potential to improve its classification performance. In the recent decade, many researchers have tried to improve LPP from different aspects, such as Discriminant LPP (DLPP) \cite{dlpp}, Orthogonal LPP (OLPP) \cite{olpp}, Parametric Regularized LPP (PRLPP) \cite{rlpp} and their extensions \cite{dlppm,2ddlpp,odlpp,udlpp}. More specifically, DLPP uses a similar approach as LDA and emphasizes preserving the local manifold structures of homogeneous data and scattering the inhomogeneous data simultaneously. On the other hand, just like LDA, DLPP also suffers the SSS problem. PRLPP regulates the LPP space in a parametric manner and extract useful discriminant information from the whole feature space rather than a reduced projection subspace of PCA. Furthermore, this parametric regularization can also be employed to other LPP based methods, such as PRDLPP, PRODLPP \cite{rlpp}. Similar to PRLPP, OLLP add an orthogonal constraint to the projection of LPP which can also be flexibly combined with other LPP methods. Different from these methods, we will try to improve LPP from its essential idea .

	LPP assumes that there exist many low-dimensional local manifolds of samples residing on the original data space. LPP intends to learn a subspace, to preserve these local manifold structures, via constructing a adjacency weight matrix which encodes the geometric information of data. This adjacency weight matrix which regarded as graph laplacian in spectral graph theory \cite{spectral}, is a discrete approximation to the Laplace-Beltrami operator $\triangledown{\mathcal{M}}$ on the manifold \cite{thesis}. Thus, the construction of this adjacency weight matrix directly determines the local manifold structure extraction. For face recognition, LPP is supervised and the entries of the weighting matrix are only determined by the distances between each two homogenous points. Therefore, LPP can only extract the local manifolds depicted by some within-class variances such as expressions and poses. Apparently, it ignores some more global variances between different persons such as facial shapes, genders, races, and face configurations, since these factors are almost invariant to the same person and corresponding to their underlying labels. We believe these kinds of information can benefit the face recognition and a natural assumption can be given that there also exist another kind of local manifold structures related to these underlying person-invariant factors. These factors are much more global than the within-class factors considered by LPP. The original space is the hybrid result of these two kinds of manifolds. Therefore, it is meaningful to learn such kind of subspace which can preserve both these two kinds of manifold structures. In this paper, we try to propose a novel method named \emph{Globality-Locality Preserving Projection} (GLPP) to address this issue. Our main contributions include:
\begin{enumerate}
  \item We propose a LPP based method to preserve the manifold structures related to both within-class variances and the person-invariant variances and attaining a more effective subspace which obtains much more classification ability than LPP in both controlled and uncontrolled environments.
  \item We formulate a 2D version of GLPP as an example to show how to combine other techniques with GLPP to develop a new GLPP-based algorithms.
\end{enumerate}

The rest of paper is organized as follows: Section 2 reviews the LPP and DLPP; Our motivation and the algorithm of GLPP and its 2D extension are described in Section 3; In section 4, several experiments are designed to demonstrate the robustness and effectiveness of GLPP; Finally, conclusion is summarized in Section 5.

\section{Related Work}
\subsection{Locality Preserving Projections (LPP)}
LPP is a linear method for face recognition and dimensionality reduction proposed by He et al \cite{lpp,lap}. In this section, we will briefly describe the model of LPP.

Given the sample set $X=[x_{1},...,x_{n}]\subset\mathcal{R}^{m}$. LPP aims at learning a projection $w$ such that it can translate the original sample space $X$ into a subspace $Y=w^{T}X=[y_{1},...,y_{n}]\subset\mathcal{R}^{d}$ which can well preserve the local manifold structures of data. This optimal projection $w$ can be solved by minimizing the following objective function
\begin{equation}\label{eq1}
\sum_{i,j}(y_{i}-y_{j})^{2}S_{ij}
\end{equation}
where the matrix $S$ is an adjacency weight matrix and $S_{ij}$ is used to measure the closeness of two points $x_{i}$ and $x_{j}$. The objective function with the choice of $S_{ij}$ will result in a heavy penalty if neighboring points $x_{i}$ and $x_{j}$ are mapped far apart. Therefore, the projection ensures that if samples $x_{i}$ and $x_{j}$ are close then their projected samples $y_{i}$ and $y_{j}$ are close as well. If LPP is used for recognition problem, it will be adopted a supervised way to construct the objective function and it can be written as follow
\begin{equation}\label{eq2}
\sum\limits_{i,j} {{{({y_i} - {y_j})}^2}{S_{ij}} = } \sum\limits_{c \in C} {\sum\limits_{i,j \in c} {{{({y_i} - {y_j})}^2}} } H_{ij}^c
\end{equation}
where the matrix $H^c$ is the adjacency matrix of the samples belonging to class $c$. Generally speaking, there are three possible ways to define the adjacency matrices $S$ or $H$:
\begin{enumerate}
  \item \textbf{Dot-product weighting}: If nodes $j$ and $i$ are connected, put $S_{ij}=x_{i}^{T}x_{j}$. Note that if $x_i$ is normalized to 1, this measurement is equivalent to the cosine similarity measure.
  \item \textbf{Heat Kernel Weighting}: If nodes $j$ and $i$ are connected, put $S_{ij}=e^{\frac{-{\vert{x_{i}-x_{j}}\vert}^{2}}{t} }$. Heat Kernel has an intrinsic connection to the Laplace Beltrami operator on differentiable functions on a manifold \cite{gnmf}.
  \item \textbf{0 - 1 Weighting}: put $S_{ij}=1$ if nodes $j$ and $i$ are connected by an edge otherwise $S_{ij}=0$. This is the simplest way to assign weights.
\end{enumerate}
 Different similarity measurements are suitable for different situations.

The objective function of LPP can be derived as:
\begin{equation}\label{eq3}
 \begin{aligned}
\sum_{i,j}(y_{i}-y_{j})^{2}S_{ij}=2(\sum_{i}w^{T}x_{i}D_{ii}x_{i}^{T}w-\sum_{i,j}w^{T}x_{i}S_{ij}x_{j}^{T}w)\\
=2(w^{T}X(D-S)X^{T}w))=2(w^{T}XLX^{T}w)
 \end{aligned}
\end{equation}
where $D$ is a diagonal matrix and its entries are column (or row, since $S$ is symmetric) sum of $S$, $D_{ii}=\sum_{j}{S_{ij}}$. Thus, $L=D-S$ is a Laplacian matrix. In the supervised case, matrix $S$ ($D$ can also be similarly denoted) is denoted as follow
\begin{equation}\label{}
S = \left( {\begin{array}{ccc}
{{H_1}}&0&0\\
0& \ddots &0\\
0&0&{{H_C}}
\end{array}} \right)
\end{equation}
where matrix $H_i$ denote the $i$th class adjacency matrix. Furthermore, there is a constraint imposed in LPP as follows
\begin{equation}\label{}
Y^{T}DY=1\rightarrow w^{T}XDX^{T}w=1
\end{equation}

Finally, this problem is reduced to find:
\begin{equation}
\hat{w}=\arg\underset{w^{T}XDX^{T}w=1}{\min}{(w^{T}XLX^{T}w)}
\end{equation}

The linear projection $\hat{w}$ that minimizes the objective functions is given by the minimum eigenvalues solution to the generalized eigenvalues problem:
\begin{equation}\label{eq5}
XLX^{T}w= \lambda {XDX^{T}w}
\end{equation}

Since the matrices $XLX^{T}$ and $XDX^{T}$ are all symmetric and positive semi-definite, the projection $\hat{w}$ which minimizes the objective function can be obtained by minimum eigenvalues solutions of the generalized eigenvalues problem. Let the column vectors $w_{1},w_{2},...,w_{d}$ to be the solutions of Equation \ref{eq5} and corresponding to the first $d$ minimum eigenvalues $\lambda_{1},\lambda_{2},...,\lambda_{d}$ . Thus, the embedding is as follows
\begin{equation}\label{eq6}
  y_{i}=W^{T}x_{i}, W=[w_{1},w_{2},...,w_{d}]
\end{equation}
where $y_{i}$ is a $d$-dimensional projected feature vector, and $W$ is the $N\times{d}$ optimized projection matrix. After obtaining the optimized projection matrix $W$, the samples can be projected via $W$ and get a much lower dimensional representation.
\subsection{Discriminant Locality Preserving Projections (DLPP)}
DLPP \cite{dlpp} is an extension of LPP borrows the idea from LDA for incorporating the discriminative information. It aims at preserving the local manifold structures of the homogenous data and scattering the adjacent classes simultaneously. The whole objective of DLPP is similar to LDA's objective function:
\begin{equation}\label{eqdlpp}
\frac{{\sum\limits_{c \in C} {\sum\limits_{i,j \in c} {{{({y_i} - {y_j})}^2}} } H_{ij}^c}}{{\sum\limits_{i,j \in c} {{{({m_i} - {m_j})}^2}} {B_{ij}}}}
\end{equation}
where $m_i$ denotes the mean sample of to the class $i$ and matrix $B_{ij}$ denotes the adjacency weight matrix of the mean samples. The definitions of the other notations are same to previous section. It is obvious that the numerator is exactly the original objective function of LPP and the denominator is the mean sample version of LPP's objective function.
Based on Equations \ref{eq3} and \ref{eq2}, the numerator of DLPP's objective function can be represented as
\begin{equation}\label{eqn}
 \begin{aligned}
 \sum\limits_{c \in C} {\sum\limits_{i,j \in c} {{{({y_i} - {y_j})}^2}} } H_{ij}^c=\sum_{i,j}(y_{i}-y_{j})^{2}S_{ij}=2(w^{T}XLX^{T}w)
 \end{aligned}
\end{equation}
Similarly, the denominator can be reduced as follow
\begin{equation}\label{eqd}
\begin{aligned}
\sum_{i,j\in{C}}(m_{i}-m_{j})^{2}B_{ij}
 =\sum_{i,j\in{C}}(w^{T}u_{i}-w^{T}u_{j})^{2}B_{ij}\\
 =2(w^{T}UGU^{T}w-w^{T}UFU^{T}w)
 =2w^{T}UKU^{T}w
 \end{aligned}
\end{equation}
The matrix $U=[u_{1},...,u_{i},...,u_{p}]\subset\mathcal{R}^{m}, i\in{C}$ denotes the mean space of samples where $u_{i}$ is the mean sample of the class $i$. And similar to the matrix $L$, $K$ is also a Laplacian matrix for measuring the weight of any two mean samples.
Substituting the Equations \ref{eqn} and \ref{eqd} in Equation \ref{eqdlpp}, the objective function of DLPP can be denoted as follow
\begin{equation}\label{}
\frac{w^{T}XLX^{T}w}{w^{T}UKU^{T}w}
\end{equation}
So, the DLPP subspace which is spanned by a set of projection vectors $w$ can be obtained by solving a programming problem:
\begin{equation}\label{}
\hat{w}=\arg\underset{w}{\min}{\frac{w^{T}XLX^{T}w}{w^{T}UKU^{T}w}}
\end{equation}

Same as LDA and LPP, this programming problem can be translated as an eigenvalue problem. Then a set of projection vectors $w$ can be achieved.

\section{Algorithm of Globality-Locality Preserving Projections}
\subsection{Motivation}
LPP is known as an efficient method for local manifold structure extraction. Many studies proved that it can perform well to the recognition problem. In this section, we focus on further improving its recognition performance. Before we introduce our improvement of LPP, several important questions should be figured out:
\begin{itemize}
  \item How does LPP preserve the local manifold structure?
  \item What kinds of local manifold structures does LPP extract in the face recognition and they are related to what kinds of information?
  \item Does any other additional manifold structures reside on the sample space and can they benefit the recognition?
\end{itemize}

\begin{figure}[h]
\setlength{\abovecaptionskip}{-0.5cm}
\setlength{\belowcaptionskip}{-0.2cm}
\centering
\includegraphics[scale=0.72]{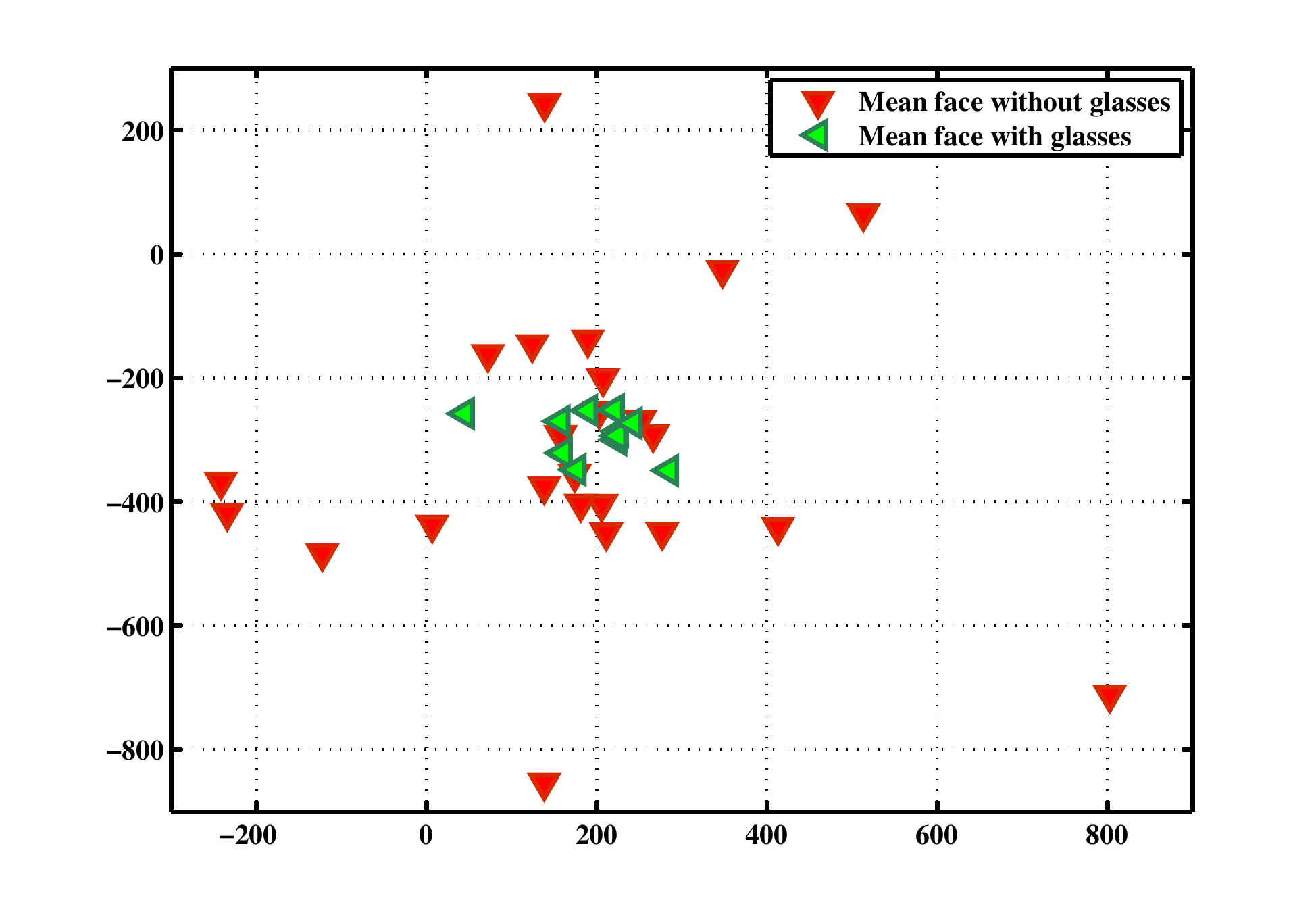}
\caption{The natural cluster in the face space. This experiment is conducted in a subset of FERET database with 35 subjects each subjects contains 6 samples ( 11 subjects wear glasses). \textbf{Note, each point in the figure represents a subject's mean face}. This 2-D subspace is learned by minimizing the globality preserving objective term of GLPP. The green points represent the mean faces with glasses and the red points represent the mean faces without glasses. \label{ttglasses}}
\end{figure}

LPP extracts the local manifold structures via providing weights based on distance between two points. The points close to each other have larger weights than the points apart. In other words, this strategy makes a very high penalty when the relative close points project far away in a learned subspace and this originates from manifold assumption \cite{lape}. Thus, this strategy allows LPP to keep the relative geometric distances between adjacent points in a learned subspace. And this geometric relationship of local adjacent points is so-called manifold (geometric) structure. LPP weights the data by the adjacency weight matrix. So, the capability and the category of the local manifold structure extraction are directly related to the adjacency weight matrix. For face recognition, LPP is supervised and the entries (weights) of adjacent weight matrix are determined by the distances between each two homogeneous samples. Therefore, LPP only extracts the local manifold structures related to the within-class variances (such as expressions and poses). More specifically, we take the heat kernel weighting as an example, the weight $w_{ij}$ of the distance between homogeneous points $i$ and $j$ can computed as $w_{ij}=e^{\frac{{-\vert{x_{i}-x_{j}}\vert}^{2}}{t} }, i,j\in{c}$ where $c$ defines the class label and $t$ is positive constant. So, the weight $w_{ij}$ is only determined by the term ${\vert{x_{i}-x_{j}}\vert}^{2},i,j\in{c}$. Let $u_c$ be the mean sample of class $c$. Thus, the term ${\vert{x_{i}-x_{j}}\vert}^{2},i,j\in{c}$ can transform into item ${\vert{x_{i}-x_{j}}\vert}^{2}={\vert{(u_c+\tilde{x}_{i})-(u_c+\tilde{x}_{j})}\vert}^{2}={\vert{\tilde{x}_{i}-\tilde{x}_{j}}\vert}^{2}$ and it exactly measures the within-class variance. Based on the previous inferences, the mean sample $u_c,c\in{C}$ is not considered in conventional LPP. This is because the mean sample which contains many meaningful information are constant to the homogenous samples, for example, the factors of facial shape, facial component shapes and skin colors which are always related to the underlying labels such as gender, races and ages, are all almost invariant to the same person. However, these person-invariant factors, which cannot be dealt by conventional LPP, can benefit the recognition. And we believe there are some manifolds corresponding to these factors. Because, there exists some natural clusters in the class level. For example, the Asians and Caucasians can be easily distinguished by skin color and the shapes of facial components. Intuitively, the Asian faces and the Caucasians faces must fall into different clusters in the whole face space. As shown in Figure \ref{ttglasses}, the mean faces with glasses clearly cluster together in a 2D-subspace learned by minimizing the globality preserving objective term of GLPP. This phenomenon also validates our assumption.

We intend to extract the manifold structure corresponding to the person-invariant factors via adding an additional objective term for constraining the LPP model to take these into consideration. This additional objective term is based on the mean sample of each class, since only the mean part is invariant to the specific person. We follow the same rule to construct its adjacency weight matrix which will be used for weighting the distance of each two mean samples. Its matrix form can expressed as follow:
\begin{equation}\label{eqg}
O_{g}=\sum_{i,j\in{C}}(m_{i}-m_{j})^{2}B_{ij}=2w^{T}UKU^{T}w
\end{equation}

\begin{figure}[h]
\setlength{\belowcaptionskip}{-0.2cm}
\centering
\subfigure[]{
\centering
\includegraphics[scale=0.325]{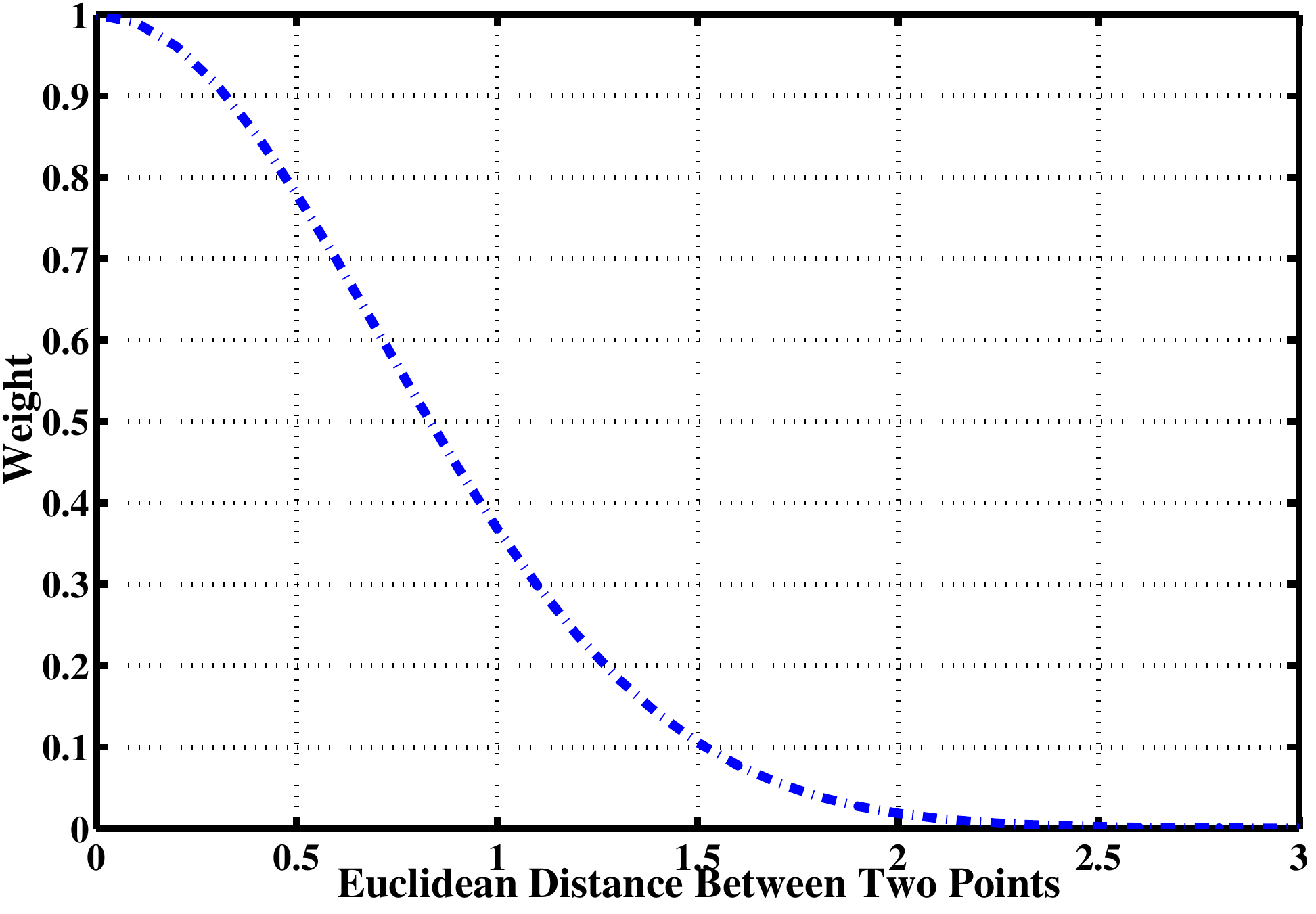}
}
\subfigure[]{
\centering
\includegraphics[scale=0.325]{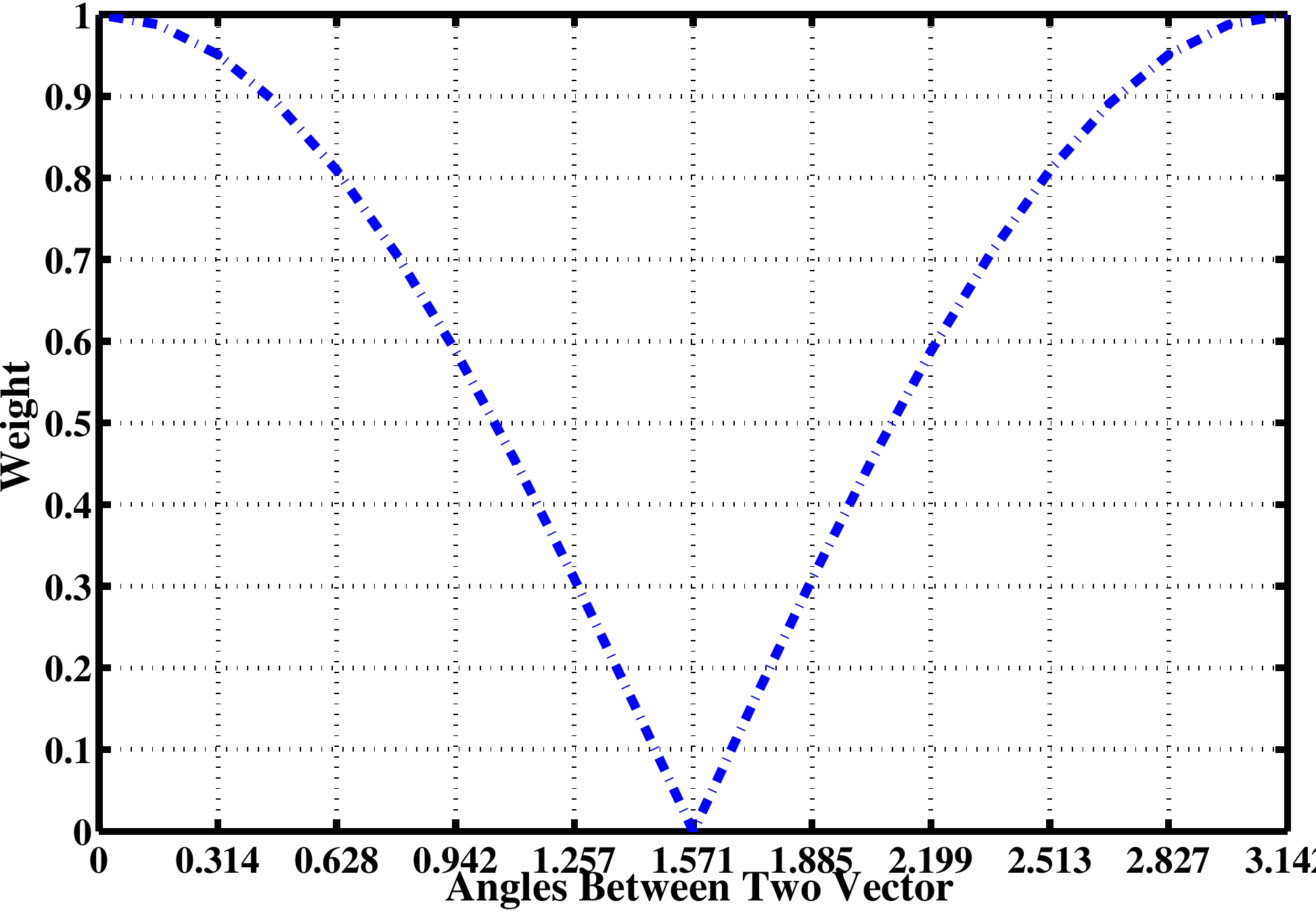}
}
\caption{The visualization of two canonical weighting mechanisms. (a) the relationship between the assigned weight and the Euclidean distance via Heat Kernel weighting. (b) the relationship between the assigned weight and the angle distance via Dot Product weighting.}
\label{weighting}
\end{figure}

We name this objective term \emph{Globality Preserving Objective Term} and its matrix form \emph{Globality Preserving Matrix} for differing to the locality preserving objective term (the original objective function of the supervised LPP,for convenience, we term it \emph{Locality Preserving Matrix} in this paper). One key point must be clarified that the globality preserving objective term actually extract the local manifold structures in the class level and it will degenerate to the unsupervised LPP when each subject only has one sample. We use the word \emph{globality} to term it, because the local manifold structures extracted by it are much more global than the ones in within-class level extracted by the locality preserving objective term. The reason why the manifolds either extracted by locality preserving objective term or gloablity preserving objective term are both local is that the weighting mechanism of LPP is nonlinear (see Figure \ref{weighting}). The weights drop sharply along with the distances (either cosine distance or Euclidean distance) increasing. Therefore, the remote points cannot effectually affect the subspace learning while the points in the local manifold play a leading role.

\begin{figure}[h]
\setlength{\belowcaptionskip}{-0.2cm}
\centering
\subfigure[]{
\centering
\includegraphics[scale=0.325]{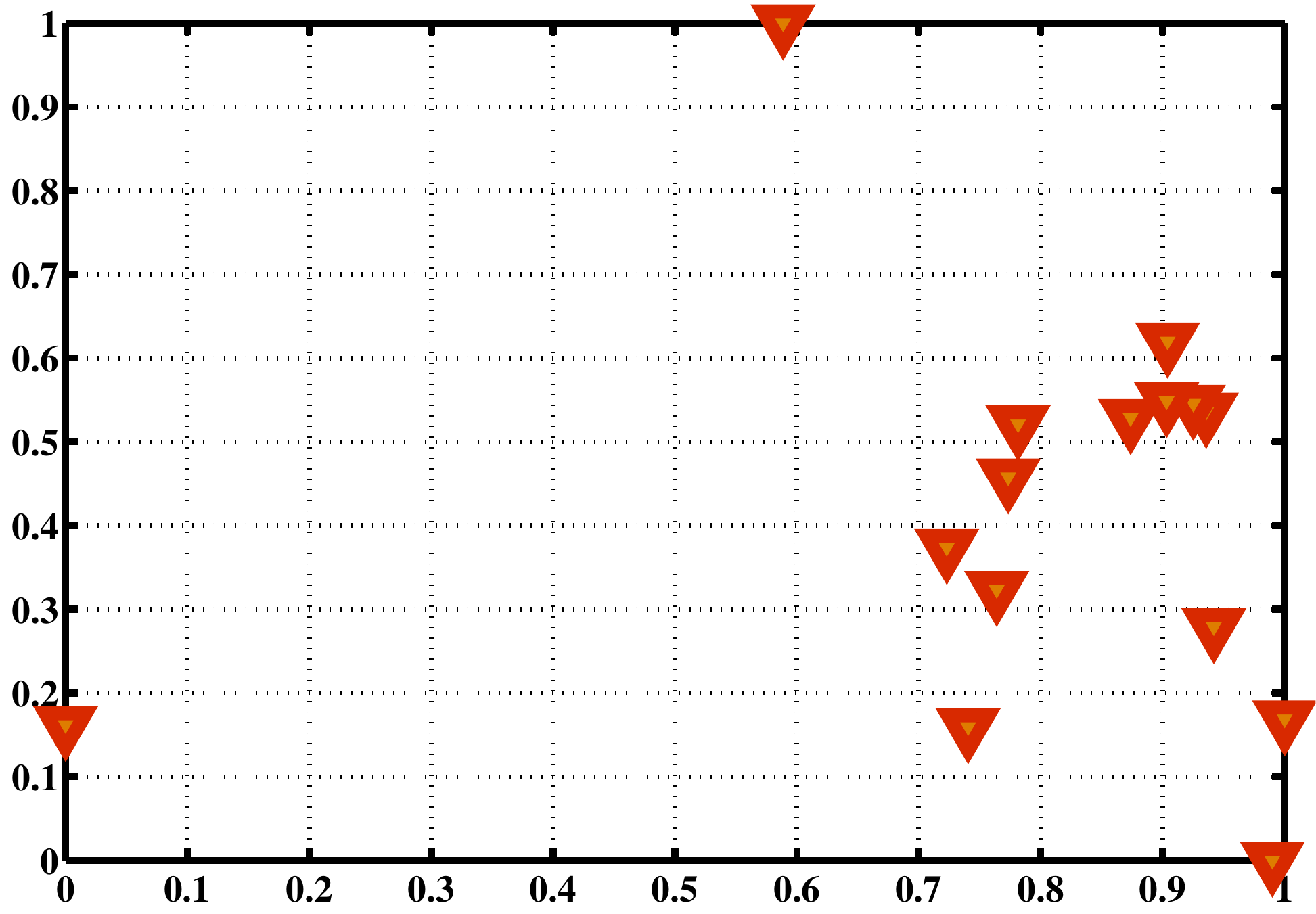}
}
\subfigure[]{
\centering
\includegraphics[scale=0.325]{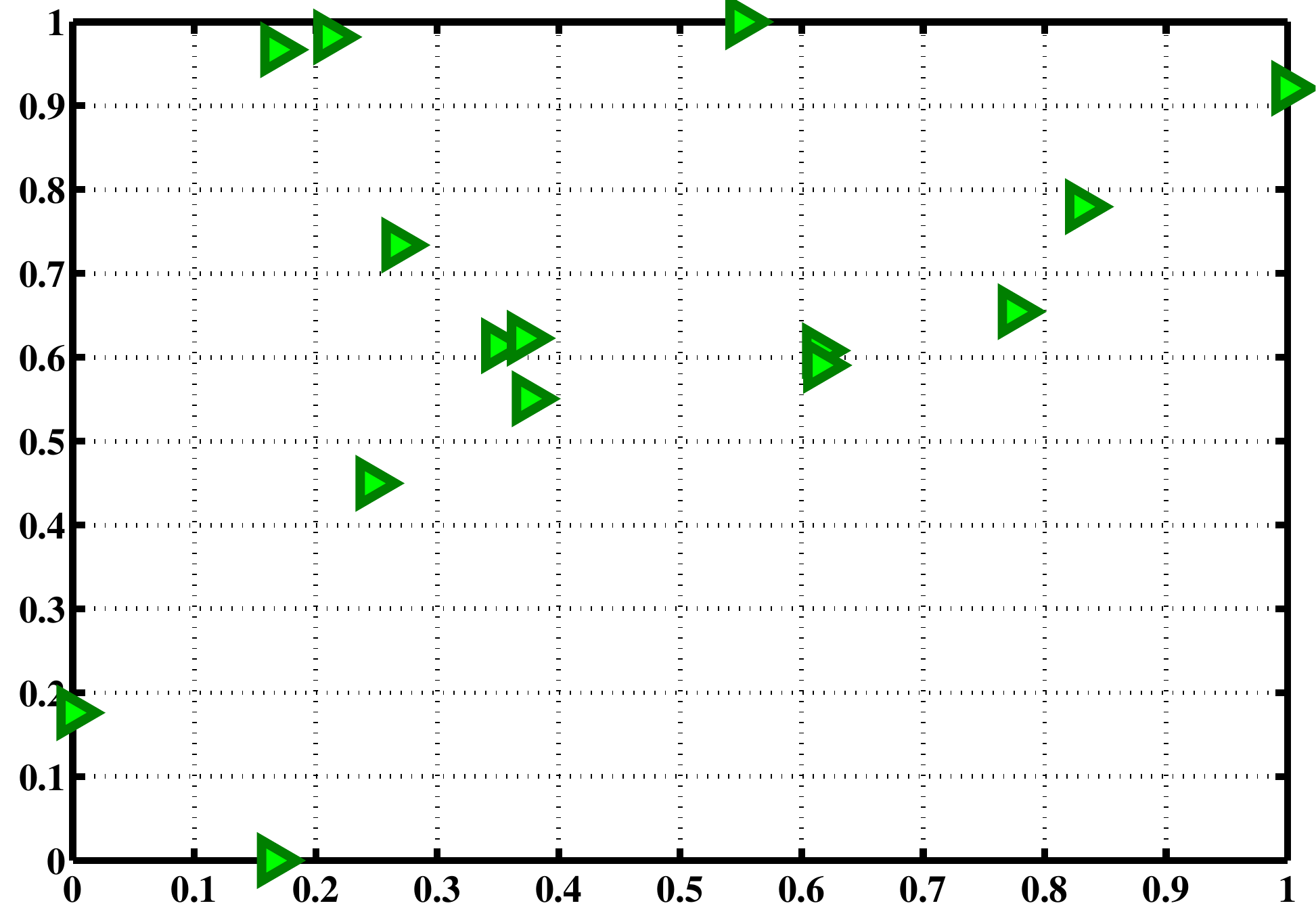}
}
\caption{The illustrations of class scattering abilities of DLPP and LDA on Yale database (15 subjects with 11 samples each). \textbf{For clarity, we just draw the mean faces in the figures.} (a)The distribution of mean faces in a subspace learned by maximizing the denominator of DLPP (the globality preserving objective term). (b)The distribution of mean faces in a subspace learned by maximizing the denominator of LDA (the between-class scatter matrix).}
\label{scattering}
\end{figure}

The globality preserving objective term is equivalent to the denominator of DLPP. The basic idea of DLPP is to scatter the nearby classes via maximizing this term and it seems to be very plausible. But, Can it really provide a \emph{good} scattering of classes? Actually, according to the interpretation from previous paragraph, this term must be locality-focused since the weighting mechanism. Maximizing this term can indeed project two local nearby classes far away but it may also lead to two remote classes project much more closer. We also conduct experiments on Yale database to illustrate the class scattering ability of DLPP. See Figure \ref{scattering}, this figure illustrates the class scattering abilities of DLPP and LDA. The right subfigure (the red points) illustrates the class scattering via maximizing the denominator of DLPP (the globality preserving matrix) and the left one illustrates the class scattering via maximizing the denominator of LDA (the between-class scatter matrix). It is clearly that the scattering performance of DLPP is not good in comparison with the classical LDA. In brief, the class scattering ability of DLPP is still questionable and DLPP breaks the natural manifold structures of the person-invariant factors in the class level.

\subsection{Globality-Locality Preserving Projections (GLPP)}\label{glppintro}
We aims at improving the recognition performance of LPP via extracting much more meaningful information from data. An additional objective term called \emph{Globality Preserving Objective Term} is added to the original objective function of LPP for preserving the \textbf{local manifold structures} corresponding to the person-invariant factors in the class level. Intuitively, this new LPP method is named as \emph{Globality-locality Preserving Projection} (GLPP).

Before we formally introduce GLPP, we should firstly define some notations. Same as the previous section, we assume matrix $X=[x_{1},...,x_{n}]\subset\mathcal{R}^{m}$ as the original sample space and the class label library as vector $C=[1,2,...,p]$. Matrix $X_{c}, c\in{C}$ is assumed to denote the subset belonging to class $c$. The matrix $U=[u_{1},...,u_{i},...,u_{p}]\subset\mathcal{R}^{m}, i\in{C}$ denotes the mean space of samples where $u_{i}$ is the mean sample of the class $i$. Matrix $M=[m_{1},...,m_{i},...,m_{p}]\subset\mathcal{R}^{d}, i\in{C}$ denotes the projected mean sample space via projecting the original mean sample space $U$ into the optimal subspace. Similarly, the projected sample space is denoted as $Y=[y_{1},...,y_{n}]\subset\mathcal{R}^{d}$. Our job is to find a projection matrix $W=[w_{1},w_{2},...,w_{d}]$ which maps the $m$-dimensional original sample space to a $d$-dimensional subspace preserving both global and local geometric structures preferably preserved subspace.

Here we give the objective function of GLPP:
\begin{equation}\label{objofglpp}
min(O_{g}+\beta{O_{w}})
\end{equation}
where the objective term $O_{g}$ denotes the gloablity preserving objective term and the objective term $O_{w}$ denotes the locality preserving objective term (the original objective function of LPP). The parameter $\beta$ is used for balancing $O_{g}$ and $O_{w}$.  A greater value of $\beta$ indicates the model pays much more attention to preserving the local manifold structures. We set $\beta{>}1$ based on a intuitive and natural assumption that the between-class variance is much greater than the within-class variance for the classification problem. These two terms are defined as follows
\begin{equation}\label{og}
O_{g}=\sum_{i,j\in{C}}(m_{i}-m_{j})^{2}B_{ij}
\end{equation}
\begin{equation}\label{ow}
O_{w}=\sum_{c\in{C}}\sum_{i,j\in{c}}(y_{i}-y_{j})^{2}S_{ij}
\end{equation}
substitute these two equations into Equation \ref{objofglpp}, then the objective can be formulated as
\begin{equation}\label{eq10}
min(\sum_{i,j\in{C}}(m_{i}-m_{j})^{2}B_{ij}+\beta{\sum_{c\in{C}}\sum_{i,j\in{c}}(y_{i}-y_{j})^{2}S_{ij}})
\end{equation}
Where matrices $S$ and $B$ is the adjacency weight matrices of the objective terms $O_{w}$ and $O_{g}$ respectively. In this paper, we choose the dot-product weighting to construct each adjacency matrix.

Finally, Equation \ref{eq10} can be manipulated by some simple algebraic steps as:
\begin{eqnarray}\label{eq11}
 & &\sum_{i,j\in{C}}(m_{i}-m_{j})^{2}B_{ij}+\beta{\sum_{c\in{C}}\sum_{i,j\in{c}}(y_{i}-y_{j})^{2}S_{ij}} \nonumber \\
 &=&\sum_{i,j\in{C}}(w^{T}u_{i}-w^{T}u_{j})^{2}B_{ij}+\beta{\sum_{c\in{C}}\sum_{i,j\in{c}}(w^{T}x_{i}-w^{T}x_{j})^{2}S_{ij}}
 \nonumber \\
 &=&2(w^{T}UGU^{T}w-w^{T}UFU^{T}w+\beta\sum_{c\in{C}}(w^{T}X_{c}D_{c}X_{c}^{T}w-w^{T}X_{c}S_{c}X_{c}^{T}w))
 \nonumber \\
 &=&2w^{T}(UKU^{T}+\beta\sum_{c\in{C}}(X_{c}L_{c}X_{c}^{T}))w
 \nonumber \\
 &=&w^{T}Aw
\end{eqnarray}
Where $K$ and $L_{c},c\in{C}$ are the Laplacian matrices and $A$ is a positive semi-definite matrix. Therefore, this problem as follow
\begin{equation}\label{eq12}
  \hat{w}=\arg\underset{w}{\min}{(w^{T}Aw)}
\end{equation}
can be transformed into a generalized eigenvalue problem (Its solving process can refer to the solving process of LPP in previous section) denoted as follow
\begin{equation}\label{eq13}
\lambda{w}=Aw
\end{equation}
The first $d$ best projections $w$ are corresponding to the first $d$ minimum nonzero eigenvalues $\lambda$. Thus we can finally obtain the GLPP projection matrix $W=[w_{1},w_{2},...,w_{d}]$. Then we can project the data into optimal subspace via GLPP projection and employ different classifiers for classification.
\subsection{A Two-Dimensional Extension of GLPP (2D-GLPP)}
In this section, we will present an algorithm termed \emph{Two-Dimensional Globality-Locality Preserving Projection} (2D-GLPP) as an example to show how to combine other techniques with GLPP to develop the new algorithm.

2D-GLPP considers the input data as an image matrix instead of a vector. Let us consider a set of images $G=[g_{1},g_{2},...,g_{N}]$ taken from an $m\times{n}$ dimensional image space. For dimensionality reduction, we should design a set of linear projections which map the original $m\times{n}$ image matrix into a ${m}$ dimensional feature space.
\begin{equation}\label{eq14}
y_{i}=g_{i}w, i=1,2,...,N
\end{equation}
where $y_{i}\in{Y=[y_{1},y_{2},...,y_{N}]}$ is the $m$ dimensional projected feature vector and $w$ is a linear projection.

Same as GLPP, we can compute the between-class and within-class adjacency matrices. However, we can not employ these Laplacian matrices directly since the input data is two dimensional. To solve this problem, the Laplacian matrices should be transformed as follows
\begin{equation}\label{eq15}
T=L\otimes{I_{m}}=
\left(                 
  \begin{array}{cccc}   
    l_{11} & l_{12} & \cdots & l_{1n}\\  
    l_{21} & l_{22} & \cdots & l_{2n}\\
    \vdots & \vdots & \ddots & \vdots\\
    l_{n1} & l_{n2} & \cdots & l_{nn}\\  
  \end{array}
\right)\otimes
\left(\underbrace{                
  \begin{array}{cccc}   
    1 & 0 & \cdots & 0\\  
    0 & 1 & \cdots & 0\\
    \vdots & \vdots & \ddots & \vdots\\
   0 & 0 & \cdots & 1\\  
  \end{array}
  }_{m}
\right)
\end{equation}
The symbol $\otimes$ is the Kronecker product of the matrices. Then, the objective function of 2D-GLPP can be expressed as follows
 \begin{equation}\label{eq16}
  \begin{aligned}
2w^{T}(M(K\otimes{I_{m}})M^{T}+\beta{\sum_{c\in{C}}}{(G_c(L_c\otimes{I_m})G_c^{T})})w\\
=2w^{T}(MZM^{T}+\beta{\sum_{c\in{C}}}{(G_cT_cG_c^{T})})w
 \end{aligned}
 \end{equation}
 Where $G_c=[g_1^{T},g_2^{T},...,g_l^{T}]\subset{G},c\in{C}$ is a $ml\times{n}$ matrix generated by arranging all the image matrices ,belong to class $c$, in column. Similarly,$M=[m_1^{T},m_2^{T},...,m_c^{T}],m_i=\sum_{j\in{c}}{g_j}$ is a $mc\times{n}$  matrix generated by arranging each class’s mean image matrix in column. Same as GLPP, this problem
 \begin{equation}\label{eq17}
 \begin{aligned}
 \hat{w}=\arg\underset{w}{\min}{(w^{T}(MZM^{T}+\beta{\sum_{c\in{C}}}{(G_cT_cG_c^{T})})w)}
 \rightarrow{\arg\underset{w}{\min}{(w^{T}Aw)}}
 \end{aligned}
 \end{equation}
 can be also finally solved as generalized eigenvalue problem.
\section{Experiments}
We evaluate the performance of the proposed GLPP and its 2D extensions on four popular face databases involving both controlled environments and uncontrolled environments. The face databases in controlled environment are ORL, FERET and Yale databases, while LFW-a is a face database in uncontrolled environment.
\subsection{Experimental Setting}
\subsubsection{Datasets}
\begin{figure}[h]
\setlength{\belowcaptionskip}{-0.6cm}
\centering
\subfigure[]{
\centering
\includegraphics[scale=0.35]{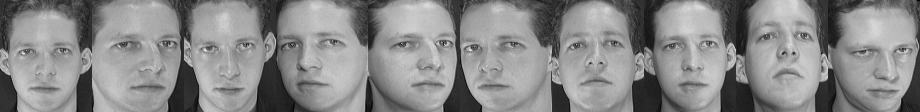}
\label{fig3a}
}
\subfigure[]{
\centering
\includegraphics[scale=0.4]{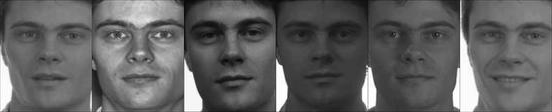}
\label{fig3b}
}
\centering
\subfigure[]{
\centering
\includegraphics[scale=0.35]{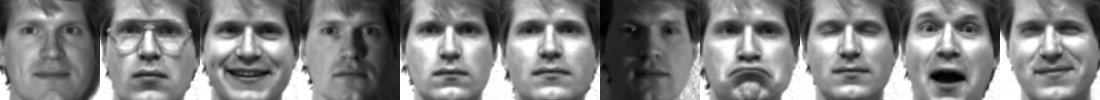}
\label{fig3c}}
\subfigure[]{
\centering
\includegraphics[scale=0.3]{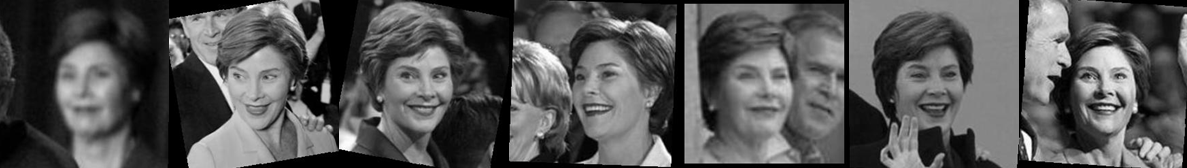}
\label{fig3d}
}
\caption{Sample face images from (a) the ORL database (b) the FERET database (c) the Yale database and (d) the LFW-a database}
\label{fig3}
\end{figure}

\begin{enumerate}
  \item The ORL database contains 400 images from 40 subjects (Figure \ref{fig3a}) \cite{orl}. Each subject has ten images acquired at different time. In this database, the subjects have varying facial expressions and facial details. And the images are also taken with a tolerance for some tilting and rotation of the face of up to $20^\circ$. For simplicity, we aligned and cropped the face image to size 32$\times$32 pixels.
  \item The FERET database contains 13539 images corresponding to 1565 subjects (Figure \ref{fig3b}) \cite{feret}. In our experiments, we select a subset which contains 436 images of 72 individuals and this subset involves variations of facial expressions, illumination and poses.
  \item The Yale database has 165 grayscale images of 15 individuals (Figure \ref{fig3c}) \cite{yale}. Every subject has different facial expressions and configurations. And the size of each image is 32$\times$32 pixels.
  \item The LFW-a database is an automatically aligned grayscale version \cite{lfwa} of the LFW database \cite{lfw} which is a database aim at studying the problem of the unconstrained face recognition (Figure \ref{fig3d}). This database is considered as one of the most challenging database since it contains 13233 images with great variations in terms of lighting, pose, age, and even image quality. We copped these images to 120$\times$120 pixels in the center, and resize them to 64$\times$64 pixels.
\end{enumerate}

\subsubsection{Compared methods and their source codes}
We compare our method with state-of-the-art methods including LDA, PCA, LPP, and DLPP. The source codes are downloaded from Prof. Deng Cai's homepage \cite{code}.
\subsection{Face Recognition}
We conducted several experiments to evaluate GLPP and compare it with PCA, LDA, LPP and DLPP in terms of recognition accuracy in controlled and uncontrolled environments. The 2D-extensions of GLPP will also be briefly evaluated in this section via comparing with 2D-PCA \cite{2dpca}, 2D-LDA \cite{2dlda} and 2D-LPP \cite{2dlpp}. We applied the nearest neighbor classifier in the Euclidean space to perform recognition. Dot product weighting is applied for constructing adjacency matrices of LPP-based methods. The recognition accuracy reported in this section is top recognition rate (the number of corrected recognized testing samples divided by the number of total testing samples). In these experiments, four cross-validation schemes include leave-one-out scheme, single-sample scheme, two-fold scheme, five-fold scheme or three-fold scheme are applied for each database according to the sample number of subjects to evaluate the performance of GLPP.
\subsubsection{Recognition Performance of GLPP in Controlled Environment}\label{controlled}
Three databases include ORL, Yale and FERET are employed in this experiment. The parameters of DLPP are deferred to the experimental section of \cite{dlpp}. And the parameter $\beta$ of GLPP is fixed to 10000. We will introduce how to learn $\beta$ in the next subsection. Before applying the face recognition methods to the databases, PCA is utilized to reduce the redundant information of data and only preserves the dimensions corresponding to nonzero eigvalues (PCAratio=1). Average Recognition Accuracy (ARA) and Standard Deviation (STD) are used to measure the recognition performance and robustness respectively.

\begin{table}[h]
\setlength{\belowcaptionskip}{-0.55cm}
  \begin{center}
    \begin{tabular}{c c c c c}
    \hline
     \multirow{2}*{Methods}
    &\multicolumn{4}{c}{Cross-Validation Schemes-Recognition Rate (ARA$\pm$STD)}\\ \cline{2-5}
    &Leave-one-out& Five-fold & Two-fold & Single samples\\
    \hline
     PCA &94.25$\pm$3.1\% &91.25$\pm$3.2\%&82.25$\pm$0.4\%&51.19$\pm$3.2\% \\
     LDA &99.00$\pm$1.7\% &98.00$\pm$1.9\%&93.25$\pm$1.8\%&47.58$\pm$4.4\%\\
     LPP &98.00$\pm$2.6\% &96.75$\pm$1.4\%&90.75$\pm$3.9\%&\textbf{54.58$\pm$4.2\%}\\
     DLPP &98.25$\pm$2.1\% &97.25$\pm$2.1\%&93.75$\pm$3.2\%&51.19$\pm$3.2\%\\
     GLPP &\textbf{99.50$\pm$1.1\%} &\textbf{98.75$\pm$1.5\%}&\textbf{96.00$\pm$1.4\%}&51.86$\pm$2.5\%\\ \hline
    \end{tabular}
    \caption{Recognition performance comparison (in percents) using ORL database }
    \label{t1}
  \end{center}
\end{table}

\begin{table}[h]
\setlength{\belowcaptionskip}{-0.55cm}
  \begin{center}
    \begin{tabular}{c c c c c}
    \hline
     \multirow{2}*{Methods}
    &\multicolumn{4}{c}{Cross-Validation Schemes-Recognition Rate (ARA$\pm$STD)}\\ \cline{2-5}
    &Leave-one-out& Five-fold & Two-fold & Single samples\\
    \hline
     PCA &89.79$\pm$18.5\% &89.33$\pm$11.9\%&88.67$\pm$2.8\%&65.70$\pm$19.4\% \\
     LDA &96.97$\pm$6.2\% &98.00$\pm$3.0\%&95.33$\pm$4.7\%& \textbf{67.27$\pm$17.0\%}\\
     LPP &99.39$\pm$2.0\% &99.33$\pm$1.5\%&96.67$\pm$4.1\%&66.67$\pm$16.7\%\\
     DLPP &99.39$\pm$2.0\% &98.67$\pm$1.8\%&97.33$\pm$3.8\%&65.70$\pm$19.4\%\\
     GLPP &\textbf{100.00$\pm$0.0\%} &\textbf{100.00$\pm$0.0\%}&\textbf{98.67$\pm$1.9\%}&66.72$\pm$17.5\%\\ \hline
    \end{tabular}
    \caption{Recognition performance comparison (in percents) using Yale database }
    \label{t2}
  \end{center}
\end{table}

\begin{table}[h]
\setlength{\belowcaptionskip}{-0.55cm}
  \begin{center}
    \begin{tabular}{c c c c c}
    \hline
     \multirow{2}*{Methods}
    &\multicolumn{4}{c}{Cross-Validation Schemes-Recognition Rate (ARA$\pm$STD)}\\ \cline{2-5}
    &Leave-one-out& Three-fold & Two-fold & Single samples\\
    \hline
     PCA &87.73$\pm$6.4\% &85.18$\pm$8.8\%&84.03$\pm$1.0\%&52.00$\pm$11.1\% \\
     LDA &94.44$\pm$4.2\% &92.36$\pm$3.2\%&90.74$\pm$3.3\%& 48.19$\pm$10.9\%\\
     LPP &94.44$\pm$3.9\% &92.82$\pm$2.8\%&92.82$\pm$2.3\%&52.08$\pm$10.1\%\\
     DLPP &95.14$\pm$4.0\% &92.36$\pm$4.2\%&90.51$\pm$4.4\%&51.99$\pm$11.1\%\\
     GLPP &\textbf{96.30$\pm$4.4\%} &\textbf{95.14$\pm$3.0\%}&\textbf{94.44$\pm$3.9\%}&\textbf{53.84$\pm$10.0\%}\\ \hline
    \end{tabular}
    \caption{Recognition performance comparison (in percents) using FERET database }
    \label{t3}
  \end{center}
\end{table}

Table \ref{t1}, \ref{t2}, \ref{t3} tabulate the recognition performances of different methods on ORL,Yale and FERET datasets respectively. The proposed GLPP algorithm outperforms other methods under different training sample numbers. Even, in the case of small sample size, our proposed method still can get the second place or even the first place among these five classical algorithms.

With regard to the robustness of methods (indicated by standard deviation), GLPP obviously outperforms LPP and DLPP on Yale and ORL databases and obtain a similar performance to the LPP on FERET database according to the observations from Table \ref{t1}, \ref{t2}, \ref{t3}.
\subsubsection{Recognition Performance of GLPP in Uncontrolled Environment}
 In LFW-a database, the sample number of every subject is different. The LFW-a database is divided into two subsets, each subject in the first subset (1100 images with 147 subjects) contains 6-10 samples while each subject in the second subset (3658 images with 127 subjects) contains more than 11 samples. We choose the first 5 samples per subject in the first subset as training samples and the rest as testing samples. Similar, the first 10 samples of each second subset's subject are used as training samples and the remaining are treated as testing samples. The parameter settings are the same as the experiments in controlled environment. Compared to the databases in controlled environment, the images on LFW database are more challenging. Therefore, 59-code LBP features \cite{lbp} are utilized as the baseline features on LFW-a database in this experiment. The block size of LBP is 16$\times$16 pixels and each block has 50\% overlap with adjacent blocks.

According to the observations of Table \ref{t4}, the experimental result demonstrates that GLPP obtains a significant improvement to LPP-family algorithms with preserving a bit more dimensions. This is because GLPP not only preserves local geometric structures related to the within-class variances, but also preserves global geometric structures related to the between-class (person-invariant) variances. The extra dimensions are the key to improve the performance. Furthermore, it is clear to see that GLPP get more gain over DLPP and LPP than the experiments in the controlled environment. This is because the LFW-a dataset contains more subjects which can help GLPP to more accurately preserve the local manifold structures of person-invariant factors and partly verify the existence of the manifolds related to person-invariant factors. Besides, the results also show that DLPP and LDA do not perform well in uncontrolled environment with comparison of their recognition performances in controlled environment. This indicates that DLPP may incorporate the characteristics of LDA and suffer similar problems of LDA.
\begin{table}[h]
\setlength{\belowcaptionskip}{-0.6cm}
  \begin{center}
    \begin{tabular}{c c c c c c}
    \hline
     \multirow{2}*{Subset}
    &\multicolumn{5}{c}{Top Recognition Rate (Dimension)}\\ \cline{2-6}
    &PCA& LDA & LPP & DLPP & GLPP\\
    \hline
     First set&27.42\%(625) &54.58\%(124)&58.29\%(144)&52.27\%(131)&\textbf{63.91\%(170)} \\
     Second set&35.62\%(556) &56.99\%(130)&59.45\%(136)& 55.34\%(196)&\textbf{65.75\%(286)}\\
     \hline
    \end{tabular}
    \caption{Recognition performance comparison (in percents) using LFW-a database }
    \label{t4}
  \end{center}
\end{table}

\subsubsection{Recognition Accuracy versus Dimension}
\begin{figure}[h]
\setlength{\belowcaptionskip}{-0.2cm}
\centering
\subfigure[]{
\centering
\includegraphics[scale=0.21]{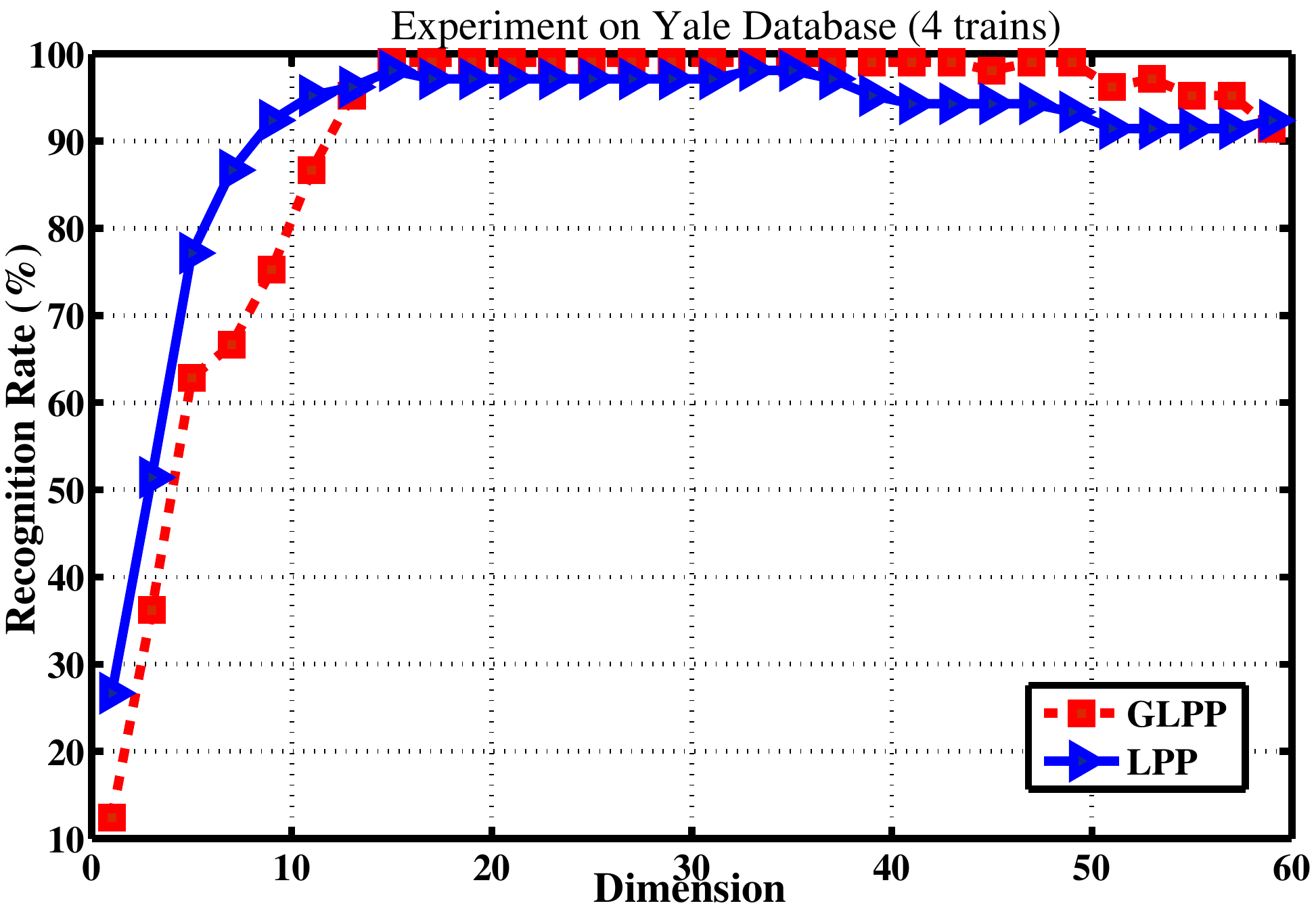}
\label{}}
\subfigure[]{
\centering
\includegraphics[scale=0.21]{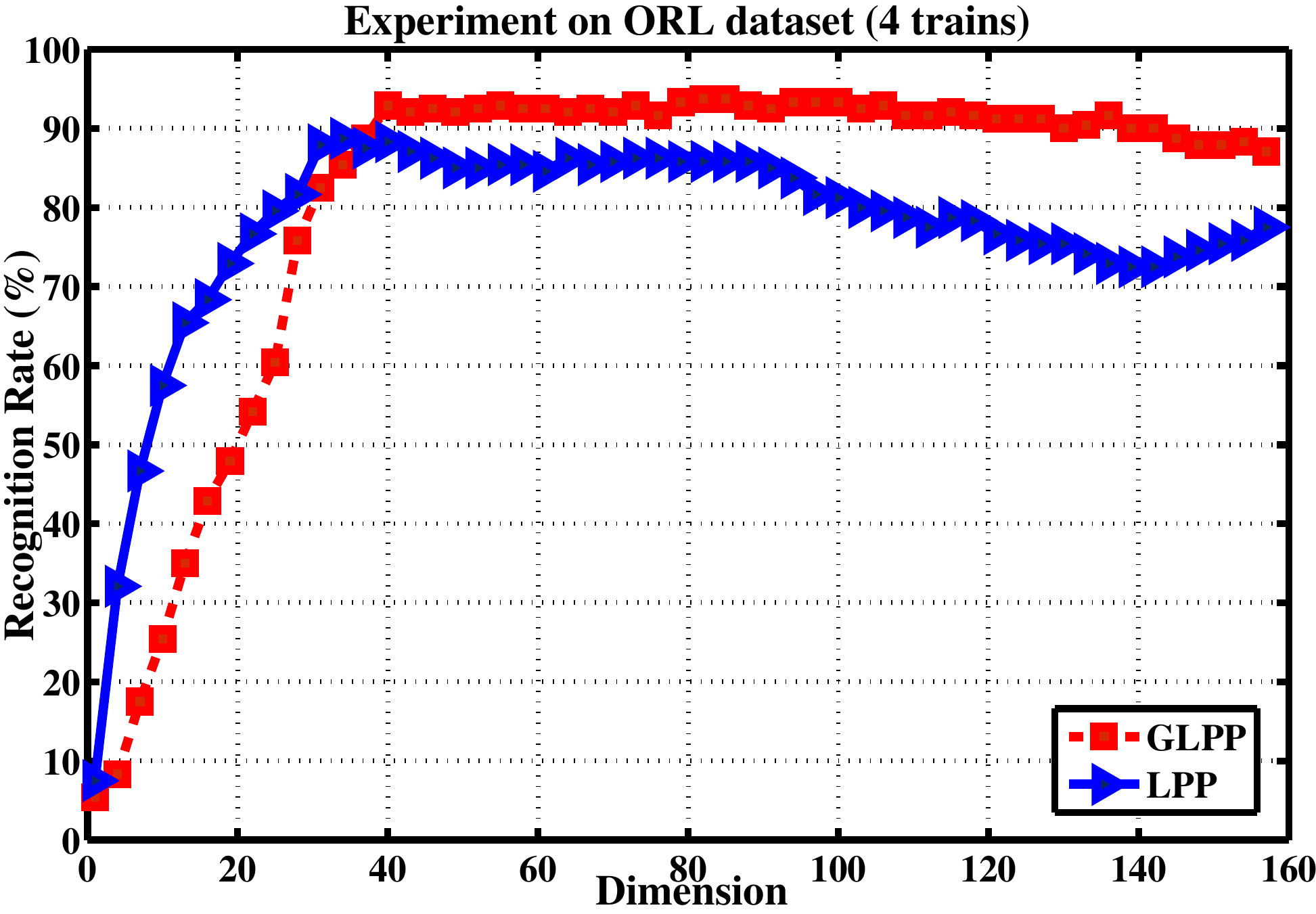}\label{}
}
\subfigure[]{
\centering
\includegraphics[scale=0.21]{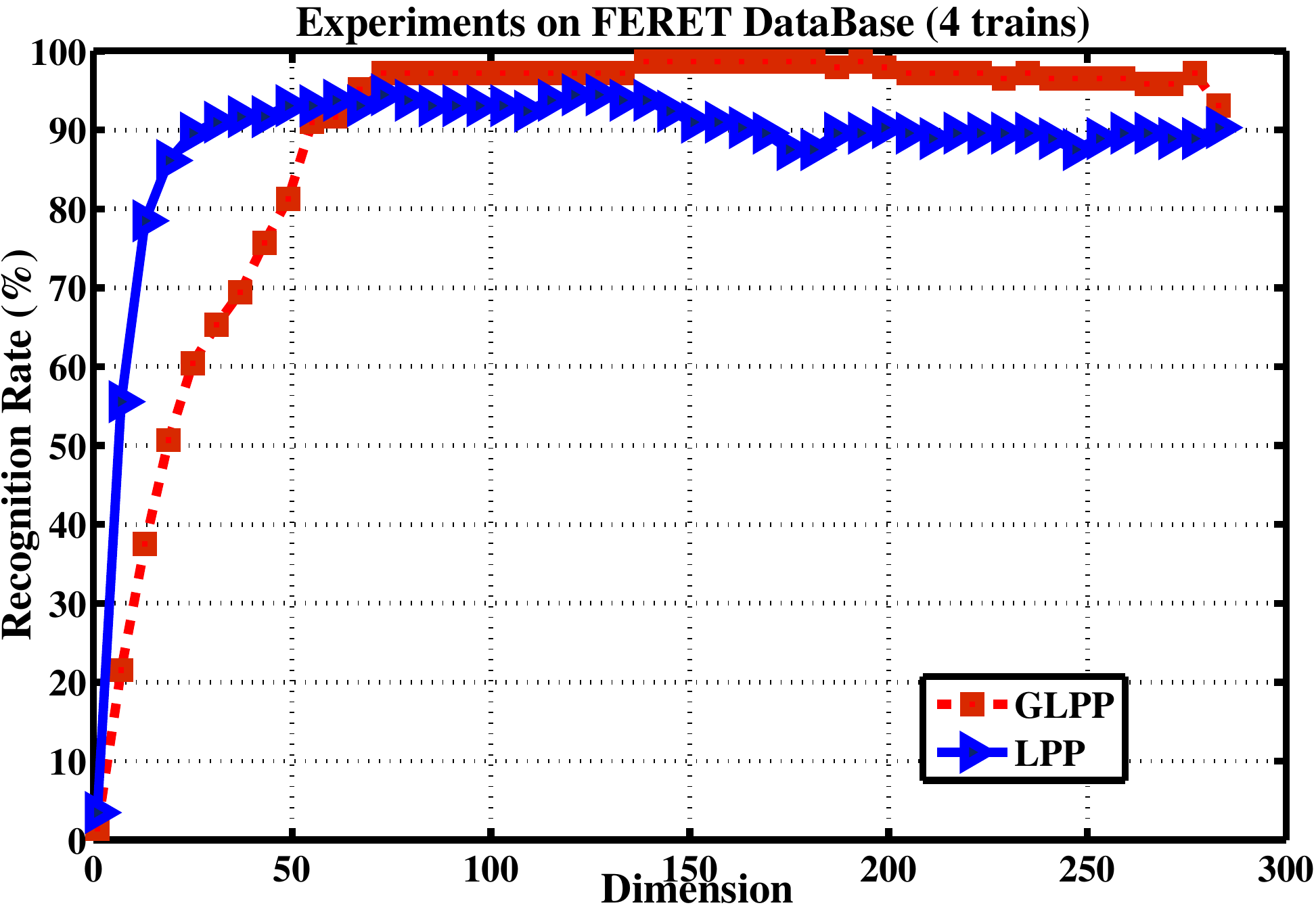}
}
\caption{the experiments of recognition accuracy versus dimension using (a) Yale database, (b) ORL database and (c) FERET database.}
\label{fig4}
\end{figure}

This experiment is conducted in the controlled environment and its experimental configuration is same to the experiment of subsection \ref{controlled}. In this experiment, first four samples of each subject are used for training and the remaining samples are used for testing. As shown in Figure \ref{fig4}, we plot the relationship of recognition accuracy and dimension. Based on the experimental results, we can see that LPP (the blue curve) obains better recognition accuracy in a relatively low dimensional space. But, GLPP (the red curve) soon outperforms LPP as the dimension increases. Moreover, the dimension corresponding to the top recognition rate of GLPP is still at a low level and acceptable for practical application.
\subsubsection{Training Time of GLPP}
We examined the training cost of GLPP and compared it with LDA, PCA and LPP. The experimental hardware configuration is CPU: 2.2 GHz, RAM: 2G. Table \ref{t5} shows the CPU time spent on the training phases by these linear methods using MATLAB. In this experiment, we select five samples of each subject for training. According to the experimental results of Table \ref{t5}, the proposed GLPP has a similar training time of the LPP.
\begin{table}
\setlength{\belowcaptionskip}{-0.55cm}
  \begin{center}
    \begin{tabular}{c c c c c}
    \hline
     \multirow{2}*{Dataset}
    &\multicolumn{4}{c}{Methods {(Seconds)}}\\ \cline{2-5}
    &PCA & LDA & LPP & GLPP\\
    \hline
     Yale &0.1248&0.1092&0.2652&0.2496 \\
     ORL &0.1404&0.1248&0.2496&0.2496\\
     FERET &2.0592&1.0452&3.3696&3.6660\\
   \hline
    \end{tabular}
    \caption{The comparison of training time (in seconds) of four linear methods}
    \label{t5}
  \end{center}
\end{table}
\subsubsection{Recognition Performance of 2D-GLPP}
This subsection is a brief introduction of the experiment of 2D-GLPP compared with 2D-LDA \cite{2dlda}, 2D-PCA \cite{2dpca}, 2D-LPP \cite{2dlpp}. Linear regression classifier is used as classifier. Three cross-validations are applied to the experiment on Yale database.

The results from Table \ref{t6} indicate 2D-GLPP perform better than other three compared methods with a smaller standard deviation.
\begin{table}
\setlength{\belowcaptionskip}{-0.55cm}
  \begin{center}
    \begin{tabular}{c c c c c}
    \hline
     \multirow{2}*{Datasets}
    &\multicolumn{4}{c}{2D linear methods-recognition rate (ARA$\pm$STD)}\\ \cline{2-5}
    &2D-PCA & 2D-LDA & 2D-LPP & 2D-GLPP\\
    \hline
     Leave-one-out &\textbf{99.39$\pm$2.0\%}&95.15$\pm$10.4\%&98.18$\pm$3.1\%&\textbf{99.39$\pm$2.0\%} \\
     Five-fold &98.67$\pm$1.8\%&94.67$\pm$6.9\%&98.67$\pm$1.8\%&\textbf{99.33$\pm$1.5\%}\\
     Two-fold &96.00$\pm$3.8\%&90.67$\pm$5.7\%&97.33$\pm$1.9\%&\textbf{99.39$\pm$2.0\%}\\
   \hline
    \end{tabular}
    \caption{Recognition performance comparison (in percents) using Yale database}
    \label{t6}
  \end{center}
\end{table}
\subsection{Learning the Parameter $\beta$}\label{tdglpp}
The parameter $\beta$ of GLPP plays an important role to trade off between the preservation of local manifolds related to within-class factors and the preservation of the local manifolds related to person-invariant factors. Therefore, it is very important to find the optimal value of $\beta$ to maximize the performance of GLPP. For obtaining this value, two experiments are applied to learn the relationship between $\beta$ and recognition performance.
\begin{figure}[h]
\setlength{\belowcaptionskip}{-0.15cm}
\centering
\subfigure[]{
\centering
\includegraphics[scale=0.21]{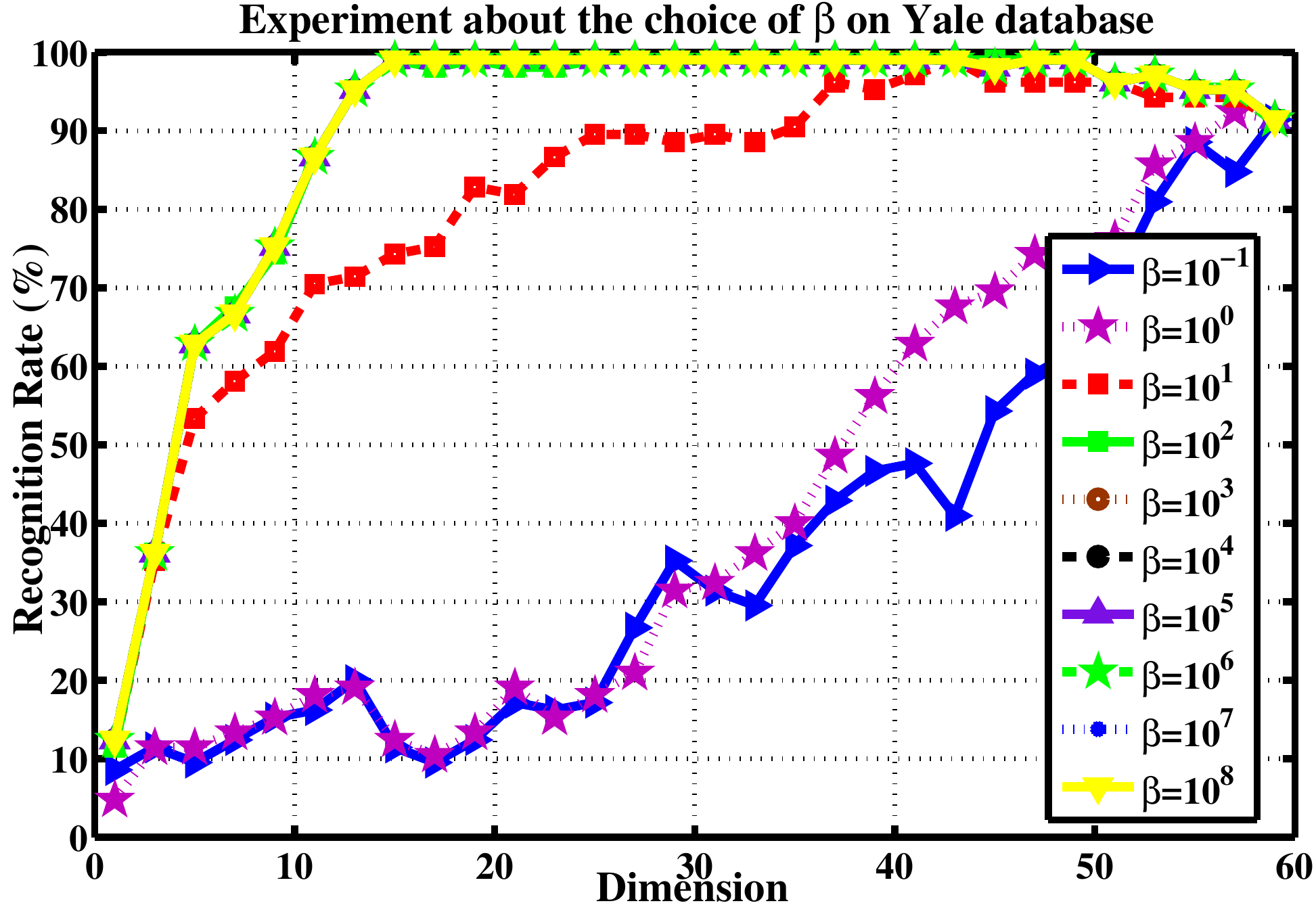}
\label{}}
\subfigure[]{
\centering
\includegraphics[scale=0.21]{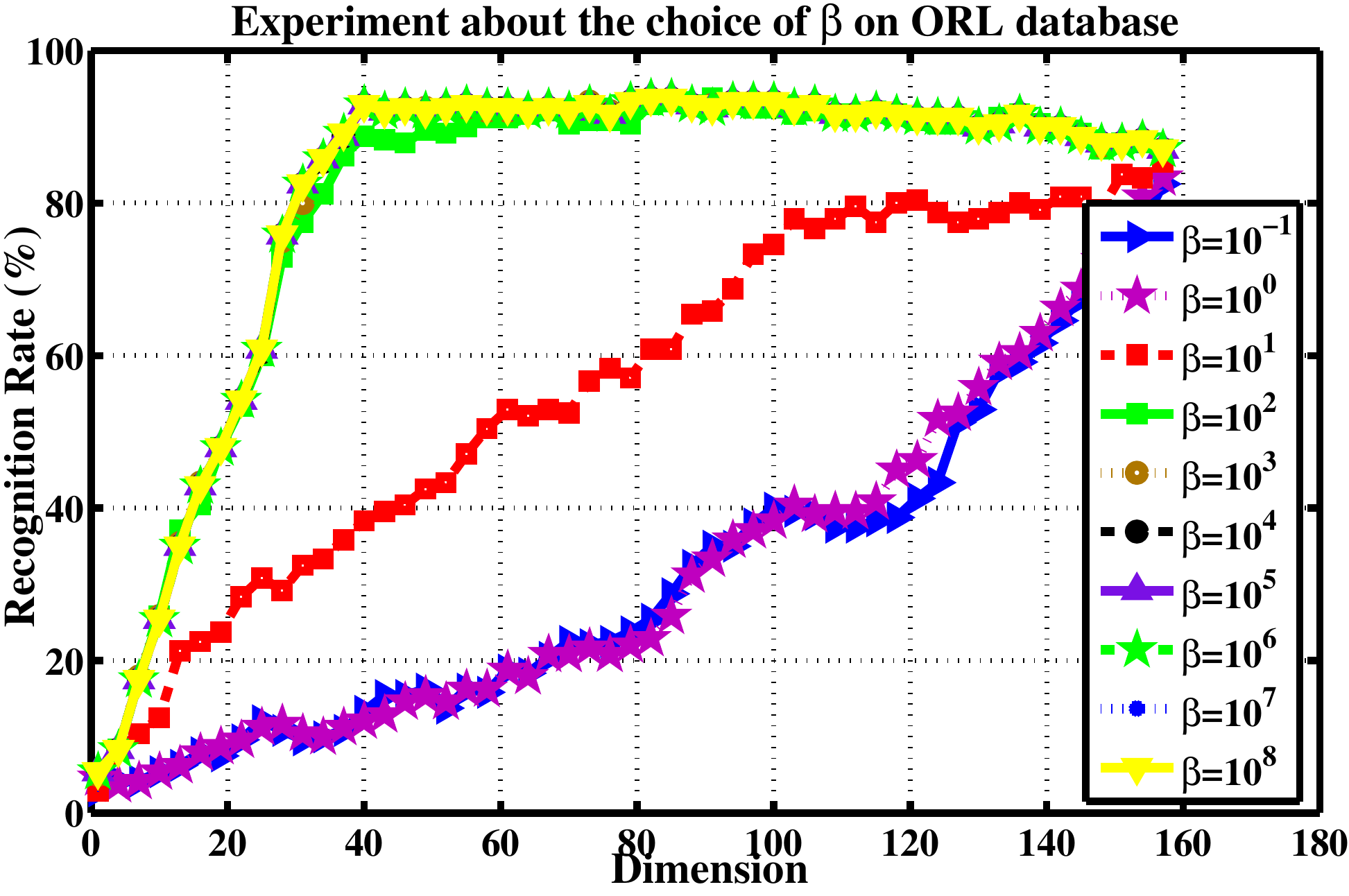}\label{}
}
\subfigure[]{
\centering
\includegraphics[scale=0.21]{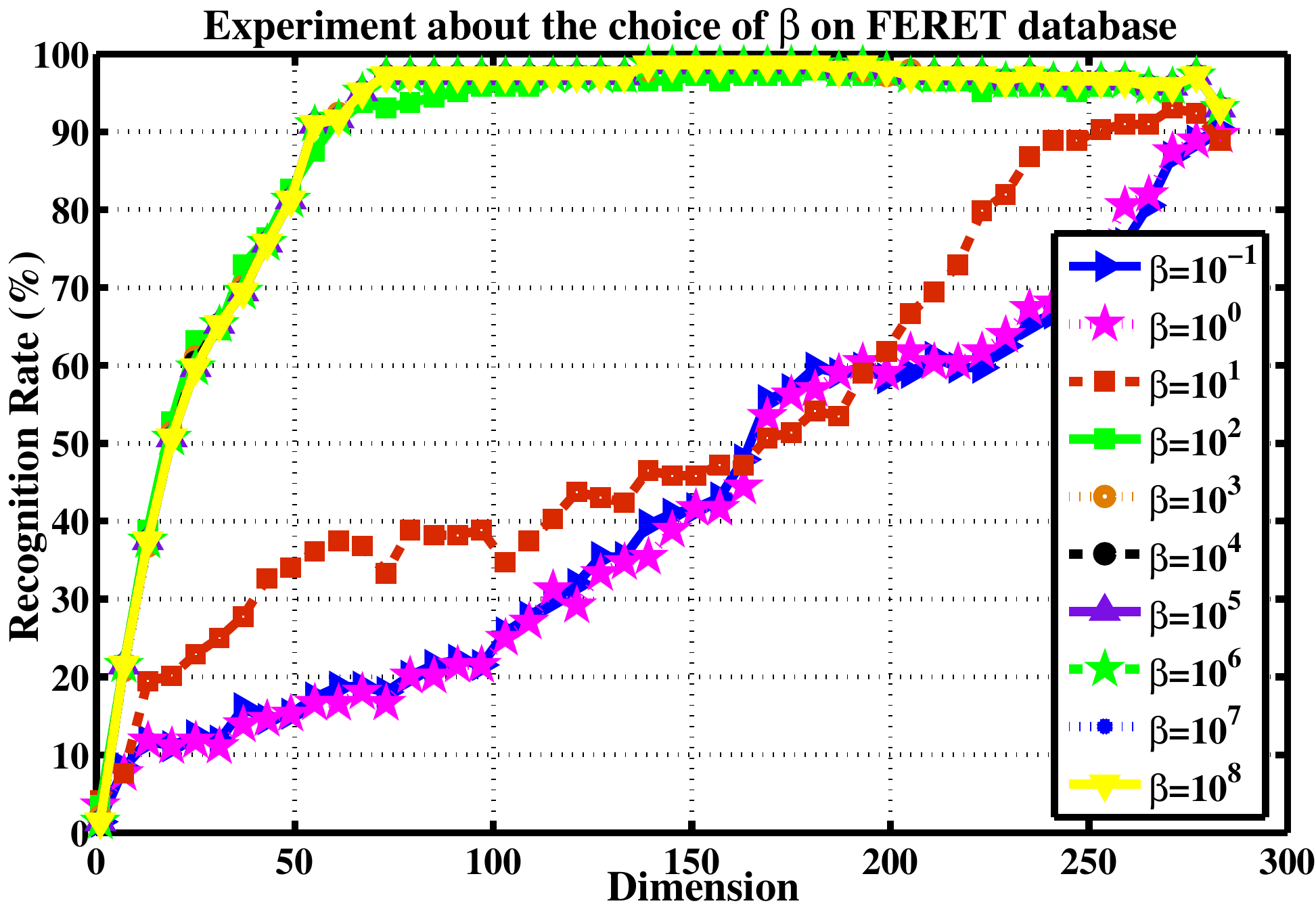}
}
\caption{Dimension vs recognition accuracy under different beta values using (a) Yale database, (b) ORL database and (c) FERET database.}
\label{fig5}
\end{figure}

\begin{figure}[h]
\setlength{\belowcaptionskip}{-0.4cm}
\centering
\subfigure[]{
\centering
\includegraphics[scale=0.21]{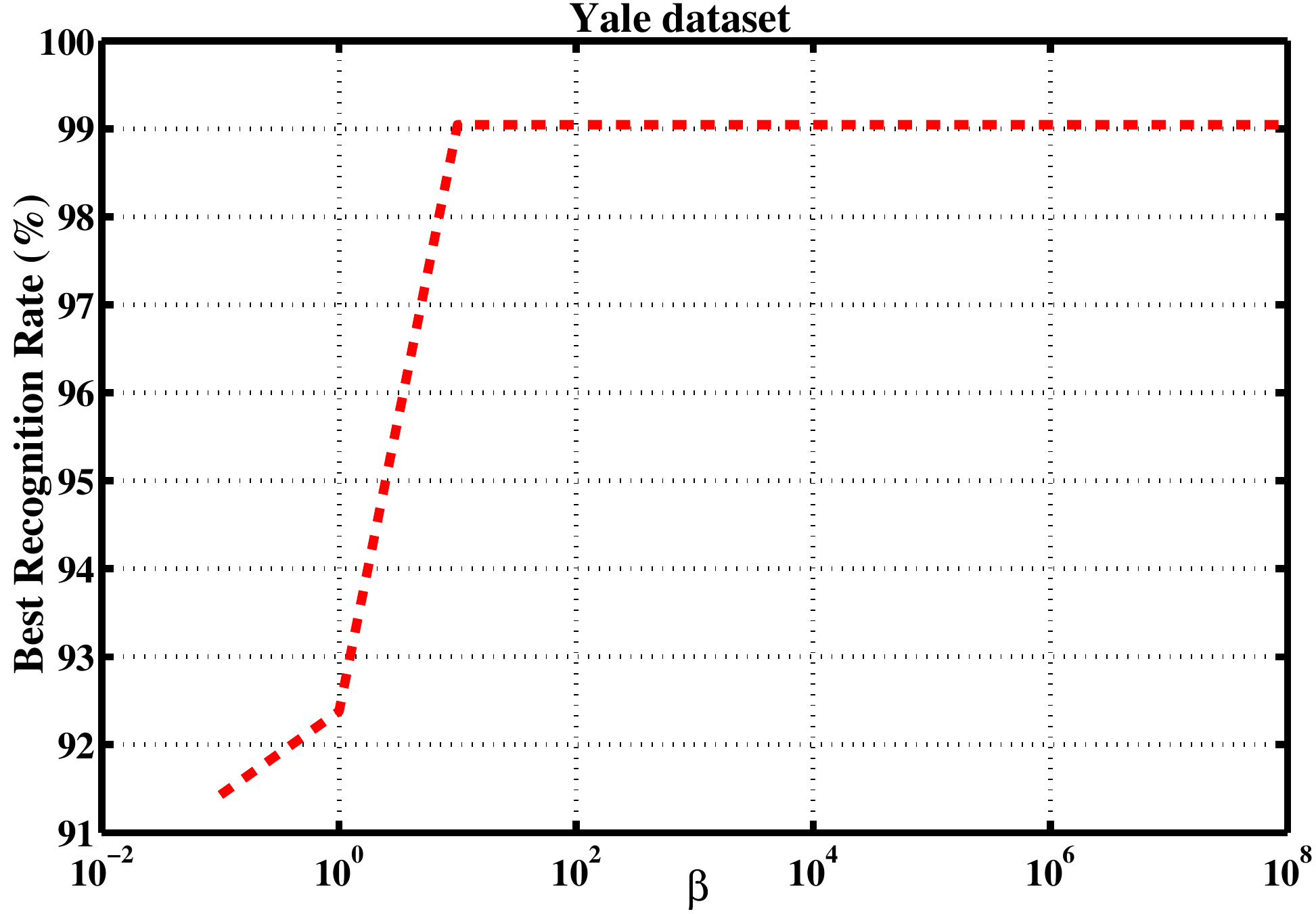}
\label{}}
\subfigure[]{
\centering
\includegraphics[scale=0.21]{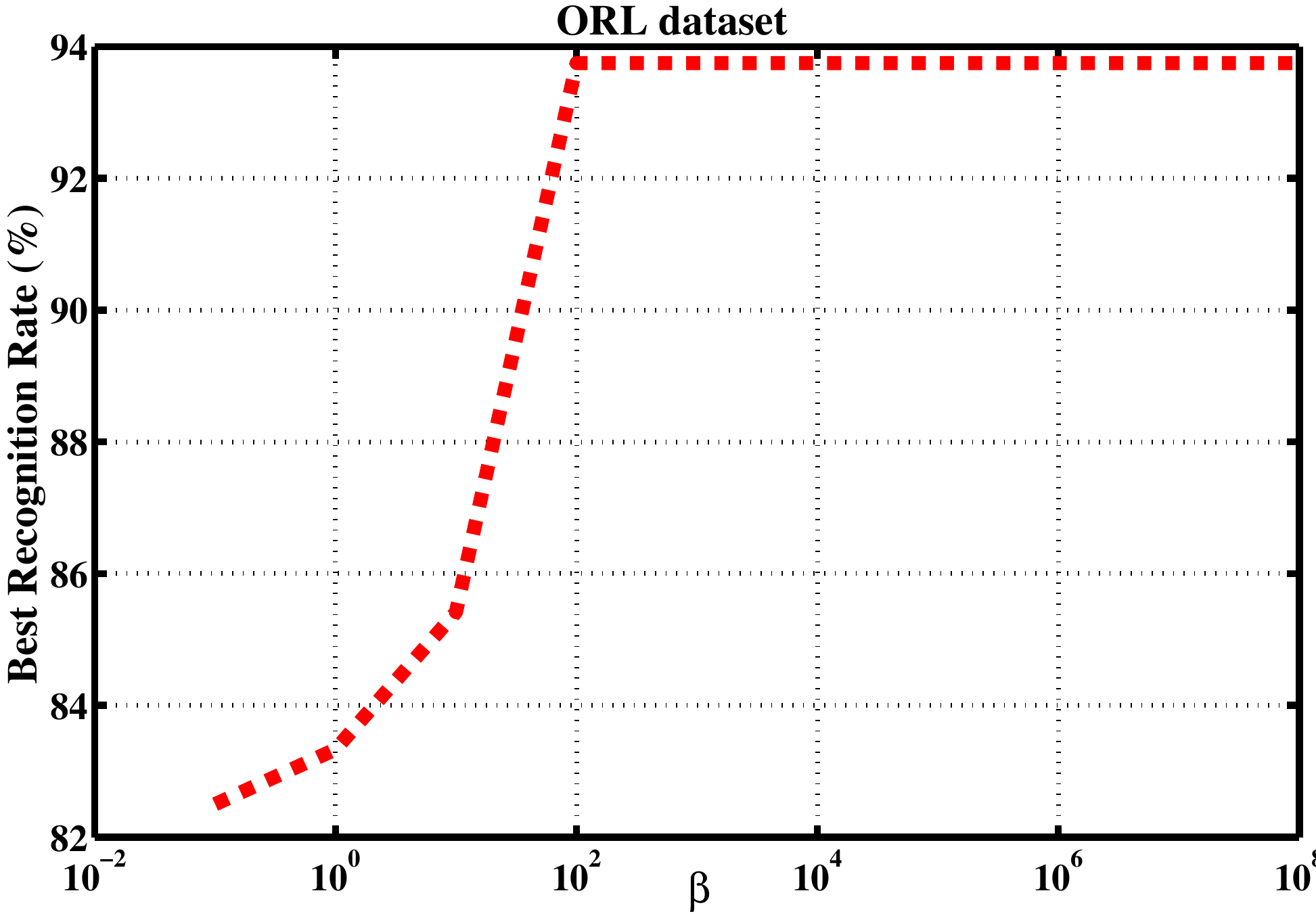}\label{}
}
\subfigure[]{
\centering
\includegraphics[scale=0.21]{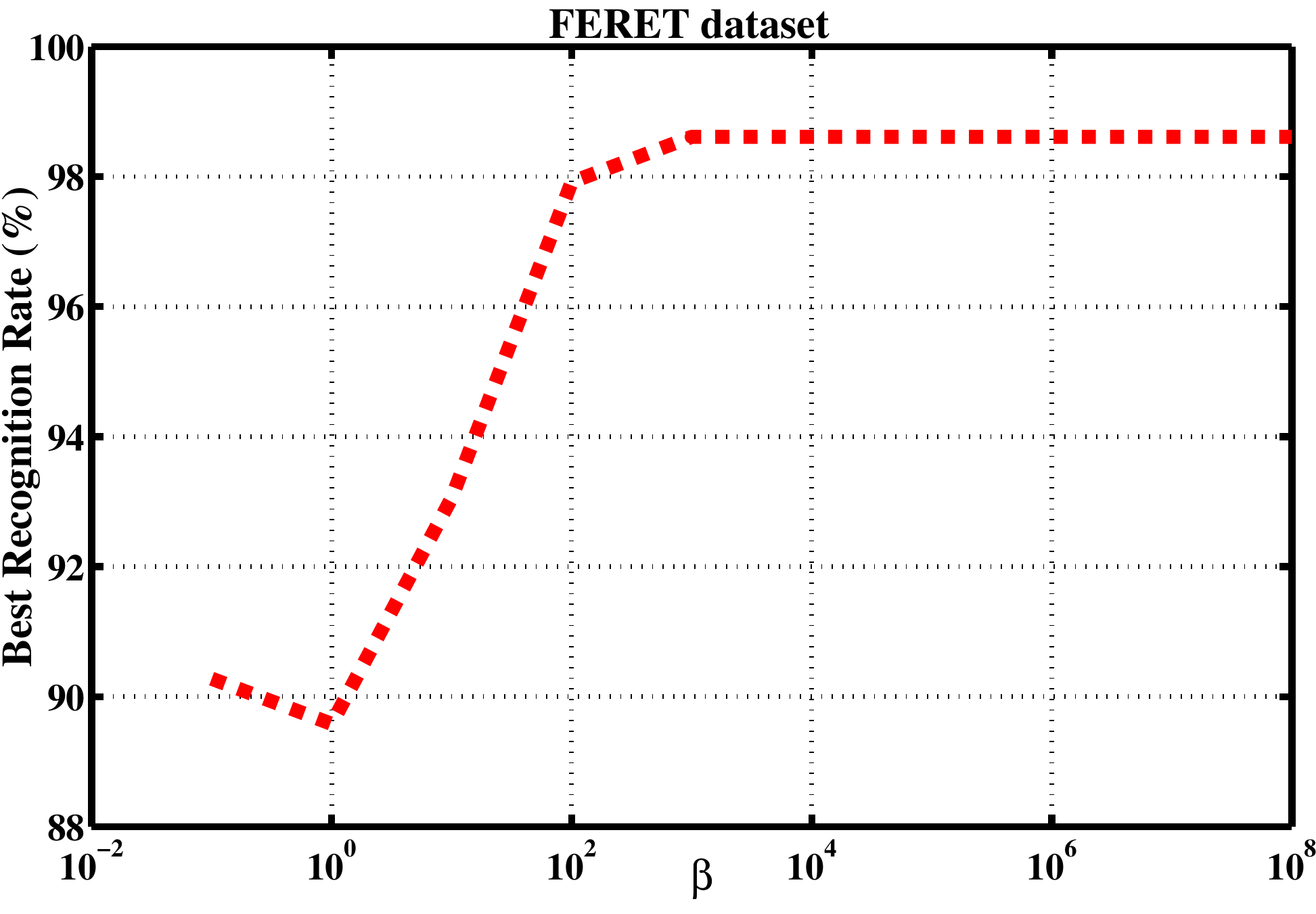}
}
\caption{The top recognition rates under different beta values using (a) Yale database, (b) ORL database and (c) FERET database.}
\label{fig6}
\end{figure}

Figure \ref{fig5} plots the curve for describing the relationships between dimension and recognition rate under different $\beta$ where the X axis indicates the preserved dimension and the Y axis indicates the recognition rate. Figure \ref{fig6} illustrates the influence of $\beta$ to the top recognition rate where X axis indicates $\beta$ and Y axis indicates the top recognition rate. Based on these curves, we can learn that GLPP achieves the best performance when $\beta$ is greater than 1000 and get the poor results when it is smaller than 1 which verifies our assumption in section \ref{glppintro}. Besides, the recognition accuracy is insensitive to $\beta$ when $\beta$ is greater than 1000. Thus we suggest that a number greater than 1000 can be assigned to $\beta$ for achieving good performance.
\subsection{Discussion}
The following observations can be made from the experimental results listed in Tables \ref{t1}-\ref{t6} and Figures \ref{fig4}-\ref{fig6}.

The proposed GLPP outperforms LDA, PCA, LPP and DLPP in both controlled and uncontrolled environments. GLPP is more suitable for extracting more meaningful information from samples and face recognition since it not only preserves the local manifold structures corresponding to within-class variances like LPP, but also can extract the person-invariant (between-class) features like PCA, and it can be evidenced from the results in Table \ref{t1}-\ref{t4} on the ORL, Yale, FERET and LFW-a database. The gains of GLPP over the best recognition accuracy of LPP are 1.5\%, 2\%, 5.25\% under leave-one-out, five-folds, two-folds cross-validation schemes respectively on ORL database. The gains of GLPP over LPP are 1.9\%, 2.3\%, 1.6\%, 1.8\% under leave-one-out, three-folds, two-folds, one sample cross-validation schemes respectively on FERET database. Particularly, the improvement of recognition performance in uncontrolled environment is much more remarkable. The gains of GLPP increase to 5.6\% and 6.3\% on the first subset of LFW-a database and the second subset of LFW-a database respectively. The reason why GLPP gets more gains over LPP in uncontrolled environment is that, GLPP can better describe and preserve the local manifold structures of person-invariant factors since the LFW-a database has more subjects.

GLPP can be extended to other manifold learning methods as LPP. In this paper, we proposed 2D-GLPP as the instance to show how to combine other techniques with GLPP to develop new algorithm. Moreover, this 2D-extension can achieve better performance compared with other classical 2D linear face recognition method includes 2D-PCA, 2D-LDA and 2D-LPP according to the experimental result in Table \ref{t6}.

The experimental results from section \ref{tdglpp} demonstrate that GLPP can achieve the best performance and its performance is not insensitive to $\beta$ when it is greater than 1000. So, $\beta$ can be fixed which can make parameter space smaller.

\section{Conclusion}
We have proposed new linear projection method for face recognition. The proposed method is designed to refine the original objective of LPP into two parts and exploit more meaningful information from samples, resulting in better recognition performance than LPP. Moreover, our proposed method appears to the first LPP based algorithm that formally incorporates the features related to invariant-person factors via simultaneously preserving both local manifold structures of within-class factors and invariant-person factors. Furthermore, our proposed GLPP can be extended to other manifold learning methods, for instance, 2D-GLPP, and similar performance improvement has been obtained. According to the development of 2D-GLPP, the proposed method is also likely to be extended to other statistical techniques such as maximum margin criterion \cite{dlppm}, orthogonal basis constraint \cite{olpp,odlpp} and parametric regularization technique \cite{rlpp}, which will be explored in our future work.
\section*{Acknowledgement}
This work described in this paper was partially supported by National Natural Science Foundations of China (NO. 0975015 and 61173131), Fundamental Research Funds for the Central Universities (No. CDJXS11181162), Key Science and Technology Project of Chongqing (No. CSTC2009AB2230). And the authors would like to thank the instructive suggestion from Prof. Ahmed Elgammal and the comments from anonymous reviewers and editors.
\label{}




\bibliographystyle{elsarticle-num}
\bibliography{mybib}

\begin{thebibliography}{10}
\expandafter\ifx\csname url\endcsname\relax
  \def\url#1{\texttt{#1}}\fi
\expandafter\ifx\csname urlprefix\endcsname\relax\def\urlprefix{URL }\fi
\expandafter\ifx\csname href\endcsname\relax
  \def\href#1#2{#2} \def\path#1{#1}\fi

\bibitem{pca}
M.~Turk, A.~Pentland, Eigenfaces for recognition, J. Cognitive Neuroscience
  3~(1) (1991) 71--86.

\bibitem{lda}
P.~N. Belhumeur, P.~Hespanha, D.~J. Kriegman, Eigenfaces vs. fisherfaces:
  Recognition using class specific linear projection, IEEE Transactions on
  Pattern Analaysis and Machine Intelligence (1997) 711--720.

\bibitem{nmf}
D.~D. Lee, H.~S. Seung, Learning the parts of objects by nonnegative matrix
  factorization, Nature 401 (1999) 788--791.

\bibitem{lpp}
X.~He, P.~Niyogi, Locality preserving projections, in: NIPS, MIT Press, 2003.

\bibitem{lap}
X.~He, S.~Yan, Y.~Hu, P.~Niyogi, H.~jiang Zhang, Face recognition using
  laplacianfaces, IEEE Transactions on Pattern Analysis and Machine
  Intelligence 27 (2005) 328--340.

\bibitem{2dpca}
J.~Yang, D.~Zhang, A.~Frangi, J.-Y. Yang, Two-dimensional pca: A new approach
  to appearance-based face representation and recognition, IEEE Transactions on
  Pattern Analysis and Machine Intelligence 26~(1) (2004) 131--137.

\bibitem{2dlda}
M.~Li, B.~Yuan, 2d-lda: A statistical linear discriminant analysis for image
  matrix, Pattern Recognition Letters 26~(5) (2005) 527--532.

\bibitem{2dlpp}
S.~Chen, H.~Zhao, M.~Kong, B.~Luo, 2d-lpp: A two-dimensional extension of
  locality preserving projections, Neurocomputing 70~(4-6) (2007) 912--921.

\bibitem{2ddlpp}
Y.~Weiwei, Two-dimensional discriminant locality preserving projections for
  face recognition, Pattern Recognition Letters 30~(15) (2009) 1378--1383.

\bibitem{isomap}
J.~B. Tenenbaum, A global geometric framework for nonlinear dimensionality
  reduction, Science 290 (2000) 2319--2323.

\bibitem{lle}
S.~T. Roweis, L.~K. Saul, Nonlinear dimensionality reduction by locally linear
  embedding, Science 290 (2000) 2323--2326.

\bibitem{kernel}
S.~K. Zhou, R.~Chellappa, B.~Moghaddam, Intra-personal kernel space for face
  recognition, in: FG, 2004, pp. 235--240.

\bibitem{kernels}
M.-H. Yang, Face recognition using kernel methods, in: NIPS, MIT Press, 2001,
  pp. 1457--1464.

\bibitem{klda}
J.~Lu, K.~Plataniotis, A.~Venetsanopoulos, Face recognition using kernel direct
  discriminant analysis algorithms, IEEE Transactions on Neural Networks 14~(1)
  (2003) 117--126.

\bibitem{cipca}
S.~Chen, T.~Sun, Class-information-incorporated principal component analysis,
  Neurocomputing 69~(1-3) (2005) 216--223.

\bibitem{dbs}
K.~Das, Z.~Nenadic, An efficient discriminant-based solution for small sample
  size problem, Pattern Recognition 42~(5) (2009) 857--866.

\bibitem{ssslda}
L.-F. Chen, H.-Y.~M. Liao, M.-T. Ko, J.-C. Lin, G.-J. Yu, A new lda-based face
  recognition system which can solve the small sample size problem, Pattern
  Recognition 33~(10) (2000) 1713--1726.

\bibitem{mmm}
H.~Li, T.~Jiang, K.~Zhang, Efficient and robust feature extraction by maximum
  margin criterion, IEEE Transactions on Neural Networks 17~(1) (2006)
  157--165.

\bibitem{mms}
F.~Song, D.~Zhang, D.~Mei, Z.~Guo, A multiple maximum scatter difference
  discriminant criterion for facial feature extraction, IEEE Transactions on
  Systems, Man, and Cybernetics, Part B 37~(6) (2007) 1599--1606.

\bibitem{tnmf}
T.~Zhang, B.~Fang, Y.~Y. Tang, G.~He, J.~Wen, Topology preserving non-negative
  matrix factorization for face recognition, IEEE Transactions on Image
  Processing 17~(4) (2008) 574--584.

\bibitem{onmf}
Z.~Li, X.~Wu, H.~Peng, Nonnegative matrix factorization on orthogonal subspace,
  Pattern Recognition Letters 31~(9) (2010) 905--911.

\bibitem{nmff}
D.~Guillamet, J.~Vitri¨¤, Non-negative matrix factorization for face
  recognition 2504 (2002) 336--344.

\bibitem{gnmf}
D.~Cai, X.~He, J.~Han, T.~S. Huang, Graph regularized non-negative matrix
  factorization for data representation, IEEE Transactions on Pattern Analysis
  and Machine Intelligence 33~(8) (2011) 1548--1560.

\bibitem{lape}
M.~Belkin, P.~Niyogi, Laplacian eigenmaps and spectral techniques for embedding
  and clustering, in: NIPS, Vol.~14, 2001, pp. 585--591.

\bibitem{ge}
S.~Yan, D.~Xu, B.~Zhang, H.~Zhang, Graph embedding: A general framework for
  dimensionality reduction, in: CVPR, 2005, pp. 830--837.

\bibitem{rlpda}
X.~Gu, W.~Gong, L.~Yang, Regularized locality preserving discriminant analysis
  for face recognition, Neurocomputing 74~(17) (2011) 3036--3042.

\bibitem{mmd}
R.~Wang, S.~Shan, X.~Chen, Q.~Dai, W.~Gao, Manifold-manifold distance and its
  application to face recognition with image sets, IEEE Transactions on Image
  Processing 21~(10) (2012) 4466--4479.

\bibitem{dlpp}
W.~Yu, X.~Teng, C.~Liu, Face recognition using discriminant locality preserving
  projections, Image and Vision Computing 24~(3) (2006) 239--248.

\bibitem{olpp}
D.~Cai, X.~He, J.~Han, H.~Zhang, Orthogonal laplacianfaces for face
  recognition, IEEE Transactions on Image Processing 15~(11) (2006) 3608--3614.

\bibitem{rlpp}
J.~Lu, Y.-P. Tan, Regularized locality preserving projections and its
  extensions for face recognition, IEEE Transactions on Systems, Man, and
  Cybernetics, Part B 40~(3) (2010) 958--963.

\bibitem{dlppm}
G.-F. Lu, Z.~Lin, Z.~Jin, Face recognition using discriminant locality
  preserving projections based on maximum margin criterion, Pattern Recognition
  43~(10) (2010) 3572--3579.

\bibitem{odlpp}
L.~Zhu, S.~Zhu, Face recognition based on orthogonal discriminant locality
  preserving projections, Neurocomputing 70~(7-9) (2007) 1543--1546.

\bibitem{udlpp}
X.~Yu, X.~Wang, Uncorrelated discriminant locality preserving projections, IEEE
  signal processing letter 15~(7-9) (2008) 361--364.

\bibitem{spectral}
F.~R.~K. Chung, Spectral graph theory, cbms regional conference series in
  mathematics (1996).

\bibitem{thesis}
M.~Belkin, Problems of learning on manifolds, phd thesis, university of chicago
  (2003).

\bibitem{orl}
F.~S. Samaria, F.~S. Samaria, A.~Harter, O.~Addenbrooke, Parameterisation of a
  stochastic model for human face identification (1994).

\bibitem{feret}
P.~J. Phillips, H.~Wechsler, J.~Huang, P.~Rauss, The {FERET} database and
  evaluation procedure for face recognition algorithms, Image and Vision
  Computing 16~(5) (1998) 295--306.

\bibitem{yale}
Y.~F. datadase, http://cvc.yale.edu/projects/yalefaces/yalefaces.html.

\bibitem{lfwa}
L.~Wolf, T.~Hassner, Y.~Taigman, Similarity scores based on background samples,
  in: ACCV, 2009, pp. 88--97.

\bibitem{lfw}
G.~B. Huang, M.~Ramesh, T.~Berg, E.~Learned-miller, Labeled faces in the wild:
  A database for studying face recognition in unconstrained environments
  (2007).

\bibitem{code}
D.~Cai, http://www.cad.zju.edu.cn/home/dengcai/data/data.html.

\bibitem{lbp}
T.~Ahonen, A.~Hadid, M.~Pietikainen, Face description with local binary
  patterns: Application to face recognition, IEEE Transactions on Pattern
  Analysis and Machine Intelligence 28~(12) (2006) 2037--2041.

\end{thebibliography}







\end{document}